%% file: main.tex
\documentclass[journal,twoside]{IEEEtran}
\usepackage{amsmath, amssymb, amsfonts}
\usepackage{url}
\usepackage{marvosym}
\usepackage{balance}
\usepackage[table,xcdraw]{xcolor}
\usepackage{cite}
\usepackage[hidelinks,hypertexnames=false,colorlinks=true,linkcolor=blue!80!black,citecolor=red!70!black,urlcolor=black]{hyperref}
\usepackage{graphicx}
\usepackage{booktabs}
\usepackage{multirow}
\usepackage{threeparttable}
\usepackage{makecell}
\usepackage{mathrsfs}
\usepackage{enumitem}
\usepackage{ulem}
\usepackage{float}
\usepackage{orcidlink}
\usepackage{amsmath}
\usepackage{array}
\usepackage{amsthm}
\usepackage{mathtools}
\usepackage{xurl}
\usepackage{dsfont}
\usepackage{cuted}
\usepackage{capt-of}

\usepackage[switch]{lineno}
\input{packages.tex}
\newcolumntype{L}[1]{>{\raggedright\arraybackslash}p{#1}}
\newcolumntype{C}[1]{>{\centering\arraybackslash}p{#1}}
\newcolumntype{R}[1]{>{\raggedleft\arraybackslash}p{#1}}

\definecolor{TYS}{rgb}{0.6, 0.8, 0.2}
\definecolor{DOcolor}{rgb}{1,0.45,0.0}
\definecolor{NAVYcolor}{rgb}{0.05,0,0.5}
\definecolor{rowlight}{RGB}{240,240,240}
\definecolor{postgreen}{RGB}{214,236,214}
\definecolor{noliyellow}{RGB}{255,246,200}
\definecolor{headblue}{RGB}{213,232,252}
\newcommand{\gc}[1]{\multicolumn{1}{>{\columncolor{rowlight}}c}{#1}}
\newcommand{\gctr}[1]{\multicolumn{1}{>{\columncolor{postgreen}}c}{#1}}

\newcommand{\settablefont}{\fontsize{6}{10}\selectfont}

\newcommand{\iei}{\textit{i.e.}}

\newcommand\clgray[1]{{\color{gray}{#1}}}

\usepackage{titlecaps}

\Addlcwords{a an and as at by for from in of on or the to with}

\begin{document}

\title{\huge
The Midas Touch for Metric Depth: Interpretable Metricization for Generalizable Depth Perception
}

\author{
    Mengtan~Zhang\,\orcidlink{0009-0003-3468-7680},
    Yu~Ma\,\orcidlink{0009-0000-8536-5444},
    Shiqi~Jiang\,\orcidlink{0009-0001-2458-865X},
    Yi~Feng\,\orcidlink{0009-0005-4885-0850},
    Wei~Ye\,\orcidlink{0000-0002-3784-7788},~\IEEEmembership{Member,~IEEE},
    Xieyuanli~Chen\,\orcidlink{0000-0003-0955-6681},~\IEEEmembership{Member,~IEEE},
    Hanli~Wang\,\orcidlink{0000-0002-9999-4871},~\IEEEmembership{Senior Member,~IEEE},\\
    Qijun~Chen\,\orcidlink{0000-0001-5644-1188},~\IEEEmembership{Senior Member,~IEEE},
    and Rui~Fan\,\orcidlink{0000-0003-2593-6596},~\IEEEmembership{Senior Member,~IEEE}
    \thanks{
    \textit{(Mengtan Zhang and Yu Ma contributed equally to this work.)}
    \textit{(Corresponding author: Rui Fan.)}
    }
    \thanks{Mengtan Zhang is with the Shanghai Research Institute for Intelligent Autonomous Systems, Tongji University, Shanghai 201210, P. R. China (e-mail: zmt200110@tongji.edu.cn).}
    \thanks{Yu Ma, Yi Feng, Shiqi Jiang, Wei Ye and Qijun Chen are with the College of {Electronic} \& Information Engineering, Tongji University, Shanghai {201804}, P. R. China (e-mail: {\{mayu2002, jiangsq0621, yew, qjchen\}@tongji.edu.cn, fengyi@ieee.org)}.}
    \thanks{Xieyuanli Chen is with the Department of Mechanical and Automation Engineering, The Chinese University of Hong Kong, Shatin, N.T., Hong Kong SAR, P. R. China (e-mail: xieyuanli.chen@nudt.edu.cn).}
    \thanks{Hanli Wang is with the College of {Electronic} and Information Engineering, the School of Computer Science and Technology, and the Key Laboratory of Embedded System and Service Computing (Ministry of Education), Tongji University, Shanghai {201804}, P. R. China (e-mail: hanliwang@tongji.edu.cn).}
    \thanks{
    {Rui Fan is with the College of Electronic and Information Engineering, Shanghai Institute of Intelligent Science and Technology, Shanghai Research Institute for Intelligent Autonomous Systems, Shanghai Key Laboratory of Intelligent Autonomous Systems, Urban  Mobility  Institute, State Key Laboratory of Autonomous Intelligent Unmanned Systems, and Frontiers Science Center for Intelligent Autonomous Systems (Ministry of Education), The Key Laboratory of Road and Traffic Engineering (Ministry of Education), Tongji University, Shanghai 201804, China, as well as with the National Key Laboratory of Human-Machine Hybrid Augmented Intelligence, Xi'an Jiaotong University, Xi'An, Shaanxi 710049, China (e-mail: rui.fan@ieee.org).}
    }
}

\maketitle

\begin{abstract}
Recent foundation models have drastically improved the generalization of monocular relative depth estimation, yet recovering accurate metric depth from their estimates for depth perception tasks remains challenging due to scale ambiguity and local metric-scale inconsistency. Existing approaches recover metric depth from sparse observations using either explicit global affine transformation algorithms or implicit data-driven depth completion networks. The former cannot accommodate spatially varying relative-to-metric relationships, while the latter are sensitive to the distribution of sparse observations and may not fully exploit generalizable relative depth priors.
To simultaneously address these issues, this study extends our previous work MTD and proposes MTDv2, a hybrid metric depth perception framework that combines explicit depth metricization with data-driven metric depth refinement.
The explicit depth metricization process performs segment-wise affine transformations to recover spatially varying metric scales while preserving the geometric structure of relative depth, followed by pixel-wise depth refinement that propagates metric observations along paths with minimal accumulated geometric variation. The data-driven depth refinement process subsequently extends the explicit single-path depth propagation to a learnable multi-path formulation while preserving its interpretability.
Extensive experiments conducted on 16 public datasets demonstrate the robust and generalizable depth scale recovery capability of MTDv2 across diverse sources of sparse observations and different depth estimation paradigms, as well as its broad applicability to downstream 3D perception tasks.
Code will be publicly available at \url{https://mias.group/MTD} upon publication.

\end{abstract}

\begin{IEEEkeywords}
Foundation model, depth perception, depth completion, scale recovery.
\end{IEEEkeywords}

\begin{figure*}[t]
\centering
\includegraphics[width=0.99\textwidth]{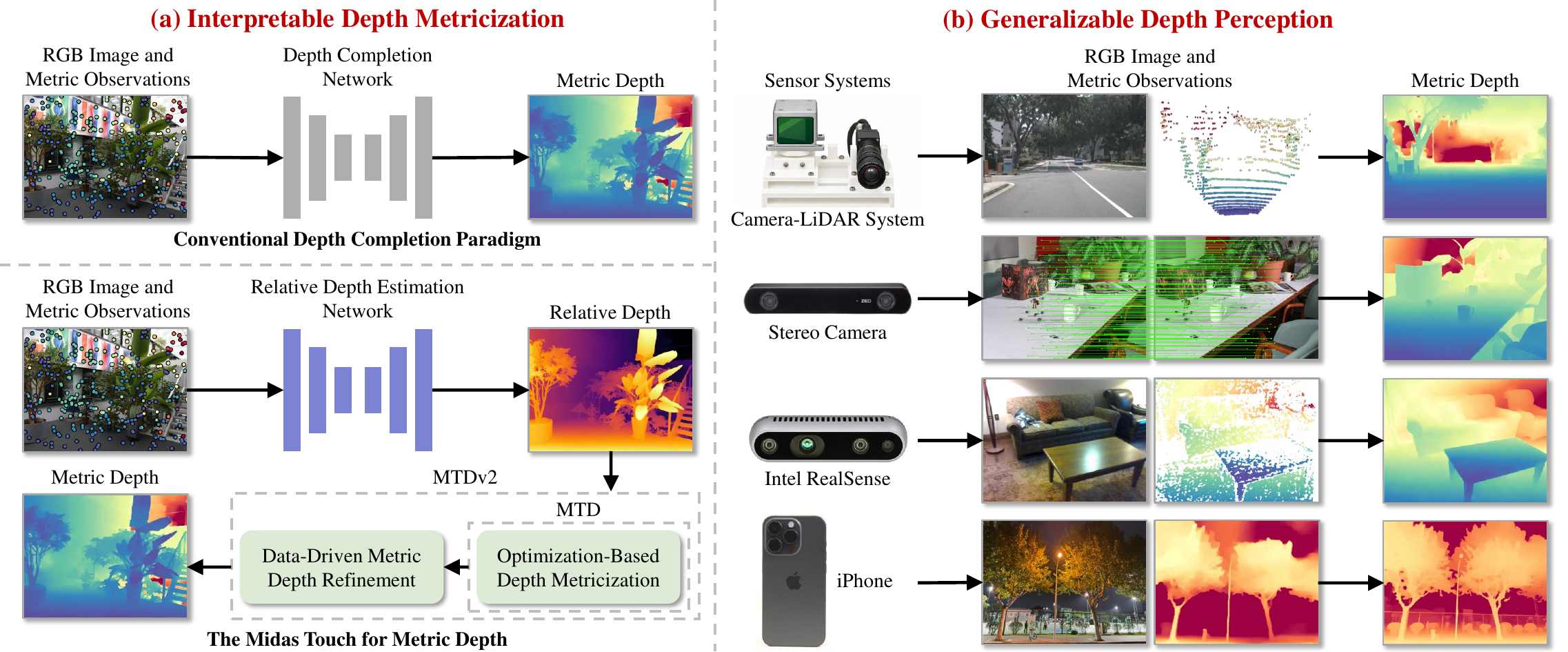}
\caption{
\textbf{Overview and capabilities of MTDv2.} By combining training-free explicit depth metricization with data-driven depth refinement, MTDv2 provides an interpretable and generalizable solution for recovering dense metric depth from sparse observations acquired using camera-LiDAR systems, stereo cameras, and other range sensors. It remains effective under extremely sparse inputs and across different sensing devices. 
}
\label{fig:fig_1}
\end{figure*}

\section{Introduction}
\label{sect.intro}

Monocular depth perception has become a fundamental component of robotics and autonomous driving~\cite{depthanythingv2}. Recent foundation models, including the MiDaS series~\cite{ranftl2020midas,birkl2023midasv3} and the Depth~Anything series~\cite{depthanything,depthanythingv2}, have significantly improved zero-shot generalization by learning rich geometric priors from large-scale data. Nonetheless, relative depth alone remains insufficient for a variety of real-world 3D applications, where metric scale and geometric consistency are indispensable~\cite{unidepth2}. This limitation is particularly pronounced in downstream tasks such as novel view synthesis (NVS) and visual odometry (VO), where visually plausible yet metrically ambiguous depth estimates can lead to inconsistent 3D reconstruction or unreliable spatial reasoning.

A straightforward approach to recovering metric scale is to align relative depth predictions with available sparse metric observations through global least-squares fitting~\cite{ranftl2020midas,depthanything,lotus}. Nevertheless, this explicit global alignment assumes that a single pair of affine transformation parameters is shared across the entire image, which is often overly restrictive~\cite{ranftl2020midas}. 
This limitation arises because different objects, surfaces, and local regions exhibit distinct relative-to-metric relationships between the relative depth predicted by foundation models trained with mixed supervision and the ground-truth metric depth~\cite{depthanythingv2}, preventing accurate metric depth recovery across the entire image.

Another line of research investigates data-driven monocular depth completion, where RGB images and sparse metric observations are jointly processed by deep neural networks~\cite{dmd3c,omnidc,ldcm, wang2024scale, shao2024nddepth}. Given sufficient training data, these methods can achieve impressive performance~\cite{lrru,dmd3c,ognidc}. Nevertheless, their generalization remains limited because neural networks may struggle to robustly encode sparse metric observations under varying metric observation densities, spatial distributions, and scene domains~\cite{bpnet,dmd3c}.
To alleviate this limitation, several studies first densify sparse observations using explicit preprocessing algorithms, such as morphological completion, interpolation~\cite{lrru}, or Poisson-style reconstruction~\cite{ldcm}, thereby providing coarse dense depth estimates for subsequent neural inference. When sparse observations are of satisfactory quality, such explicit preprocessing algorithms can produce more informative inputs and facilitate depth completion. However, under challenging conditions where metric observations are extremely sparse, highly noisy, or nonuniformly distributed, these explicit preprocessing algorithms may fail to produce geometrically reliable or sufficiently informative coarse depth estimates~\cite{bpnet,dmd3c}. 
Feeding such unreliable coarse depth into the network may introduce misleading guidance rather than useful geometric cues. Consequently, the network must compensate for preprocessing errors during training, which increases the learning burden and reduces robustness when the test-time scene domain differs substantially from that of the training data~\cite{lrru,bpnet,dmd3c,cformer}.

This motivates the development of a standalone explicit metricization algorithm that provides an interpretable formulation with broad applicability. 
Following this direction, a preliminary version of this work was presented at CVPR 2026, introducing \textbf{\textit{the Midas Touch for Depth}} (MTD)~\cite{mtd_cvpr}, a training-free framework capable of converting relative depth into metric depth using sparse observations.
MTD estimates a pair of affine transformation parameters within each image segment, propagates these parameters over a sparse segment graph, and performs discontinuity-aware pixel-level depth refinement through dynamic programming. This explicit design naturally accommodates diverse forms of sparse observations, including LiDAR points, disparities, and correspondence matches, without requiring task-specific training. 

Nonetheless, the training-free nature of MTD also introduces a practical limitation on achievable accuracy.
Within the explicit depth metricization process, imperfect image segmentation can lead to large cross-segment depth errors. In addition, handcrafted operators, manually tuned hyperparameters, and predefined propagation rules may produce suboptimal depth propagation results and leave residual depth errors, particularly in complex scenarios.
While these limitations do not diminish the value of the explicit algorithm, they motivate leveraging the strengths of deep learning to model local depth variations in complex scenarios. Importantly, such capabilities should be incorporated in a complementary manner that does not compromise the interpretability of the explicit formulation.

Therefore, this article extends our previous work into \textbf{\textit{the Midas Touch for Depth V2}} (MTDv2), a hybrid depth perception framework that combines explicit optimization-based depth metricization with data-driven metric depth refinement. 
The optimization-based depth metricization process largely follows MTD to explicitly resolve the relative-to-metric ambiguity, while making further improvements by incorporating depth information into local region segmentation, producing segments that are more geometrically and texturally coherent.
This design improves the consistency of the independently estimated affine transformation parameters, thereby reducing the burden on the subsequent data-driven component to correct large cross-segment depth errors and allowing it to focus on fine-grained depth refinement.
Subsequently, the explicitly metricized depth map serves as the initialization for data-driven depth refinement and provides dense geometric cues for the depth propagation head, complementing the original sparse observations. The geometry-guided depth propagation head predicts residual corrections to the geodesic costs and propagated depth estimates, while simultaneously producing propagation logits.
These three outputs are jointly used to formulate an implicit multi-path depth propagation mechanism. Consequently, the predicted residual corrections improve local geometric fidelity and smoothness, while the multi-path depth propagation process remains consistent with the interpretable formulation of explicit optimization. MTDv2 therefore retains the flexibility and interpretability of the explicit metricization in MTD while further benefiting from the accuracy and adaptability of data-driven metric depth refinement.

To validate the effectiveness and real-world applicability of the proposed MTDv2 framework, this article conducts extensive experiments on diverse depth perception benchmarks and downstream tasks. The experimental results demonstrate that MTDv2 achieves state-of-the-art (SoTA) zero-shot depth completion performance across different sources of sparse observations, including LiDAR points, RGB-D camera measurements, and correspondence matches, as well as different levels of observation sparsity, with particularly strong advantages when the metric observations are extremely sparse. In addition, MTDv2 is compatible with different depth estimation paradigms, enabling generalizable metric scale recovery. Finally, experiments on two representative downstream tasks, NVS and VO, further demonstrate the broad applicability of MTDv2.

As a journal version of our preliminary work~\cite{mtd_cvpr}, this study makes substantial methodological and experimental extensions, with the additional contributions summarized as follows:
\begin{enumerate}
\item MTDv2, a unified hybrid framework that extends our previous work, MTD, by integrating explicit depth metricization with data-driven metric depth refinement, thereby leveraging the complementary strengths of both paradigms for depth perception.

\item An enhanced optimization-based depth metricization process that extends MTD by incorporating complementary geometric and textural cues into local region segmentation, reducing cross-segment depth errors and the burden on subsequent data-driven refinement.

\item A data-driven depth refinement algorithm featuring a geometry-guided depth propagation head derived from the discontinuity-aware geodesic principle of MTD, enabling learnable yet interpretable local depth refinement to improve geometric fidelity and smoothness.

\item Expanded experimental evaluations across diverse datasets, sparse observation sources, observation densities, and depth estimation paradigms, demonstrating improvements of MTDv2 over MTD in accuracy, robustness, and generalizability.
\end{enumerate}

The remainder of this article is organized as follows: Sect.~\ref{sect.related_work} reviews SoTA methods for depth prediction and depth completion. Sect.~\ref{sect.method} introduces the proposed MTDv2. Sect.~\ref{sect.experiments} reports extensive experimental results, demonstrating the superiority of the proposed method and its applicability to downstream tasks. Finally, Sect.~\ref{sect.conclusion} concludes this article and discusses future research directions.

\section{Related Work}
\label{sect.related_work}

\subsection{Depth Prediction}
\label{sect.related_depth_prediction}

Foundation models have greatly improved the zero-shot generalization of monocular depth prediction~\cite{depthanythingv2,moge2}. Representative approaches, including MiDaS~\cite{ranftl2020midas,birkl2023midasv3} and Depth Anything~\cite{depthanything,depthanythingv2}, typically employ inverse-depth parameterization together with affine-invariant losses to learn relative depth from large-scale datasets with diverse depth distributions. Diffusion-based approaches~\cite{marigold,lotus,depthmaster,geowizard} further improve local details by incorporating generative priors into dense depth prediction, albeit at the expense of higher computational cost. Recent studies, such as Pixel-Perfect-Depth~\cite{pixel_perfect_depth}, exploit additional geometric priors to further enhance structural consistency. Despite these advances, existing foundation models primarily estimate relative rather than metric depth, limiting their applicability to tasks that require absolute scale, geometric consistency, or direct fusion with sparse metric observations.

Monocular metric depth prediction has also attracted considerable attention in recent years~\cite{yin2023metric3d,hu2024metric3dv2,piccinelli2024unidepth,unidepth2}. These methods aim to predict metric depth directly from a single image and have achieved impressive performance through large-scale supervised training. More recently, relative-to-metric depth transfer has emerged as an alternative paradigm. For example, TR2M~\cite{tr2m} learns pixel-wise rescaling maps to convert relative depth predictions into metric depth. Nevertheless, compared with relative depth prediction, metric depth prediction remains substantially more sensitive to scene distribution, camera configuration, and training data coverage~\cite{birkl2023midasv3, depthanythingv2}. The proposed MTDv2 complements them by preserving their strong cross-domain geometric priors while explicitly recovering metric scale from sparse metric observations.

\subsection{Depth Completion}
\label{sect.related_depth_completion}

Depth completion approaches generate dense metric depth from a single image based on sparse metric observations~\cite{lrru,cformer,nlspn,dyspn,wang2024scale,shao2024nddepth}. Data-driven approaches have achieved compelling performance by combining sparse observations with image features and refining dense predictions through learnable seed propagation~\cite{nlspn,dyspn,bpnet,dmd3c}. Early methods are primarily evaluated on in-domain benchmarks with LiDAR-projected depth maps~\cite{spn1,spn2}, whereas more recent studies~\cite{omnidc, ognidc} increasingly emphasize generalization across diverse scenes, varying observation densities, and challenging conditions with noisy or extremely sparse metric observations. For instance, OGNI-DC~\cite{ognidc} introduces optimization-guided neural iterations to couple gradient prediction with depth integration, while OMNI-DC~\cite{omnidc} further improves robustness through multi-resolution integration and large-scale training with synthetically generated sparse metric observations. Moreover, PriorDA~\cite{priorda} leverages monocular geometric priors for depth completion, whereas InfiniDepth~\cite{infinidepth} models depth as a continuous representation to enable arbitrary-resolution depth completion. Nevertheless, the generalizability of these methods still depends, to some extent, on how sparse metric observations are preprocessed before or during neural inference. This limitation motivates the incorporation of explicit geometric priors or a front-end module, particularly when metric observations are extremely sparse, noisy, or nonuniformly distributed.

A variety of approaches employ non-parametric preprocessing algorithms to densify sparse metric observations before neural inference. For example, LRRU~\cite{lrru} adopts traditional interpolation and morphological operations to construct an initial dense representation, while LDCM~\cite{ldcm} uses a Poisson-based reconstruction strategy together with depth cues from a foundation model to generate coarse geometric priors. Such preprocessing algorithms are effective because they alleviate the ambiguity associated with directly processing extremely sparse metric observations. However, the resulting depth estimates are often coarse and can become unreliable when the metric observations are extremely sparse, noisy, or nonuniformly distributed~\cite{priorda}. Therefore, the subsequent network must compensate for both preprocessing errors and local structural artifacts, increasing the learning burden and reducing robustness.

Instead of developing a purely learning-based monocular depth completion network, our previously proposed MTD~\cite{mtd_cvpr} introduces an explicit metricization algorithm that converts the relative depth predicted by foundation models into metric depth based on sparse observations, without requiring task-specific training. This design provides the framework with strong flexibility, interpretability, and robustness across diverse types of metric observations. In this extended version, the explicitly metricized results can further serve as reliable geometric priors for data-driven depth refinement. As a result, MTDv2 supports both training-free metric depth completion and learning-based depth refinement, bridging explicit metricization and data-driven depth refinement within a unified framework.

\begin{figure*}[!t]
	\centering
	\includegraphics[width=0.99\textwidth]{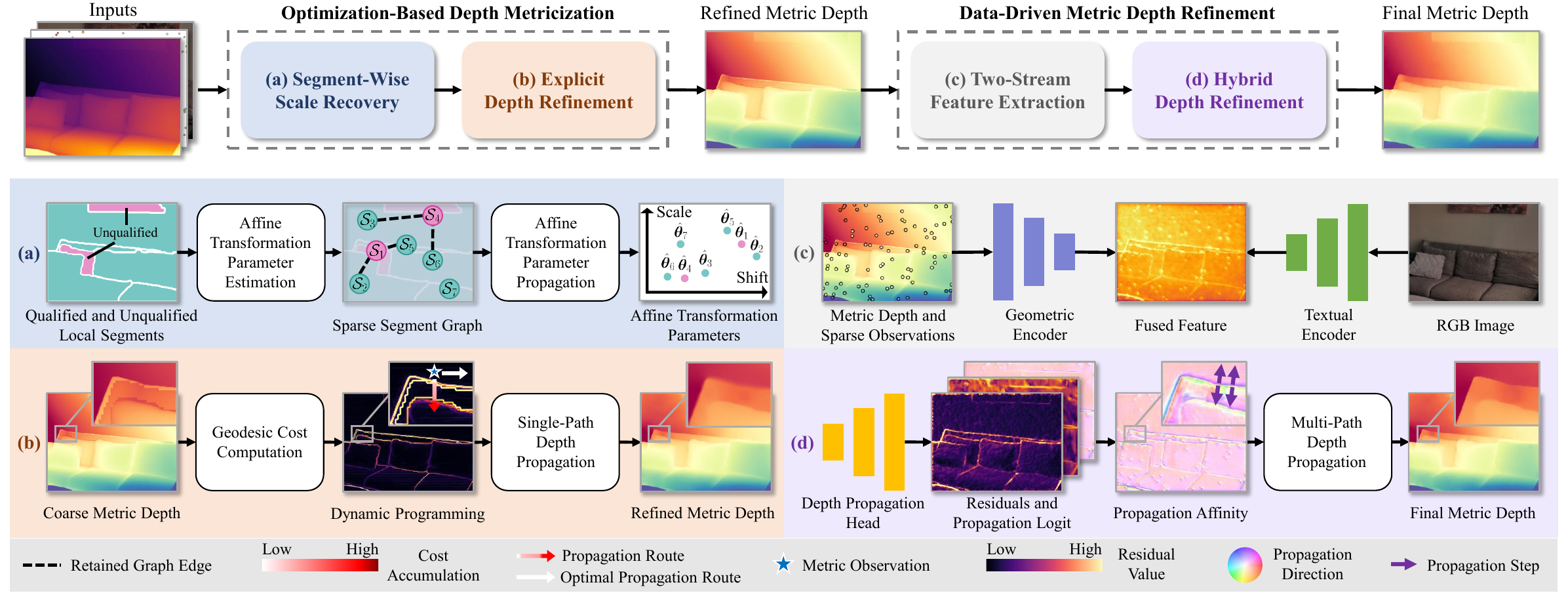}
	\caption{
    \textbf{Framework overview of MTDv2.}
    Given an RGB image, a relative depth map, and sparse metric observations, the optimization-based depth metricization is first performed.
    The affine transformation parameters are estimated for the qualified local segments and subsequently propagated to the unqualified segments via sparse graph optimization to generate a coarse metric depth map, which is then refined through single-path depth propagation based on dynamic programming, yielding the explicitly refined metric depth map.
    In the subsequent data-driven metric depth refinement process, the geometric features extracted from the explicitly refined metric depth and sparse metric observations are fused with textural features extracted from the RGB image and fed into a geometry-guided depth propagation head for learnable multi-path depth propagation, producing the final metric depth map.}
\label{fig.architectureure}
\end{figure*}

\section{Methodology}
\label{sect.method}

\subsection{Framework Overview}
\label{sect.overview}

This article presents MTDv2, a hybrid framework (see Fig.~\ref{fig.architectureure}) that combines explicit metricization with data-driven depth refinement. The framework takes an RGB image and a sparse observation map as inputs. The sparse metric observations can originate from different sources, such as LiDAR points, depth measurements, and correspondence matches, and are typically projected onto the image or associated with the corresponding pixels when necessary. The RGB image is divided into local segments based on complementary textural and geometric cues.
Within the explicit metricization process, segment-wise scale recovery is first performed by estimating independent scale and shift parameters of an affine transformation, denoted by $a$ and $b$, respectively, for local regions containing sufficient metric observations and propagating these parameters through a sparse segment graph, as detailed in Sect.~\ref{sect.segment}.
The resulting coarse metric depth is subsequently refined at the pixel level through discontinuity-aware dynamic programming, which determines the optimal depth propagation route with minimal accumulated geometric variation, as presented in Sect.~\ref{sect.geodesic}. Building upon this explicit metricization process, data-driven metric depth refinement is further performed using a geometry-guided depth propagation head that translates the same discontinuity-aware principle into a learnable local propagation mechanism, as introduced in Sect.~\ref{sect.training_guided}.

For subsequent mathematical analysis, a depth map is represented as a scalar field $z:\Omega\to\mathbb{R}$ over the image plane $\Omega\subset\mathbb{R}^{2}$, where $z(\boldsymbol{p})$ denotes the depth value at pixel $\boldsymbol{p}=(u,v)^{\top}\in\Omega$. In particular, $z_{\mathrm{rel}}:\Omega\to\mathbb{R}$ denotes the relative depth predicted by the foundation model, $z_{\mathrm{opt}}:\Omega\to\mathbb{R}$ denotes the refined metric depth obtained through optimization-based depth metricization, and $z_{\mathrm{hybrid}}:\Omega\to\mathbb{R}$ denotes the final metric depth after data-driven metric depth refinement within the complete hybrid framework.

\subsection{Scale Recovery via Sparse Graph Optimization}
\label{sect.segment}
Relative depth predicted by foundation models~\cite{depthanything, depthanythingv2} preserves informative scene geometry but lacks a reliable metric scale. When converting relative depth into metric depth based on sparse metric observations, a single global transformation is often insufficient because the relative-to-metric relationship may vary spatially across different regions of the scene. MTD tackles this problem by segmenting the image into numerous texturally consistent regions and independently estimating local affine transformation parameters for region-wise depth metricization. The effectiveness of this approach therefore depends on whether regions sharing the same affine transformation parameters are sufficiently geometrically consistent. Nevertheless, purely texture-based image segmentation may produce fragmented regions even on geometrically continuous surfaces. Since the affine transformation parameters are optimized independently for each segment, such fragmentation introduces unnecessary parameter variations and may lead to inconsistent affine transformations among adjacent regions despite their naturally similar relative-to-metric relationships. Conversely, adjacent regions that are geometrically discontinuous but exhibit insufficient textural differences may be incorrectly merged into a single segment, causing them to share the same affine transformation and consequently resulting in significant depth metricization errors.

Therefore, MTDv2 extends the purely texture-based segmentation used in MTD~\cite{mtd_cvpr} by incorporating relative depth as an additional geometric cue and employing the superpixel segmentation algorithm~\cite{felzen} to divide the image into local segments $\{\mathcal{S}_i\}_{i=1}^N$ that are consistent in both appearance and relative depth, where $N$ denotes the total number of local segments. The resulting local segments provide more reliable support for local scale recovery and subsequent graph propagation. The index set of local segments containing sufficient valid sparse metric observations, referred to as qualified local segments, is denoted by $\mathcal{Q}\subseteq\{1, \ldots, N\}$. For each qualified local segment $\mathcal{S}_i$ with $i \in \mathcal{Q}$, its affine transformation parameters are obtained by fitting the relative depth to the sparse metric observations within $\mathcal{S}_i$, expressed as follows:
\begin{equation}
\hat{\boldsymbol{\theta}}_i
=
\arg\min_{\boldsymbol{\theta}}
\sum_{\boldsymbol{p}\in\mathcal{S}_i}
m(\boldsymbol{p})
\left(
a\cdot z_{\mathrm{rel}}(\boldsymbol{p})
+b
-o(\boldsymbol{p})
\right)^2 .
\end{equation}
where $\boldsymbol{\theta}=(a,b)^\top$ stores the scale and shift parameters, $\hat{\boldsymbol{\theta}}_i=(\hat{a}_i,\hat{b}_i)^\top$ denotes the optimal parameters for the local segment $\mathcal{S}_i$, $m:\Omega\to\{0,1\}$ is an indicator function for valid sparse metric observations, with $m(\boldsymbol{p})=1$ if an observation is available at $\boldsymbol{p}$ and $m(\boldsymbol{p})=0$ otherwise, and the function $o:\Omega\to\mathbb{R}$ maps each pixel coordinate to its corresponding sparse observation value. Subsequently, for each unqualified local segment $\mathcal{S}_j$ with $j\notin\mathcal{Q}$, whose affine transformation parameters cannot be directly estimated from sparse metric observations, MTDv2 recovers its parameters by propagating affine transformations from qualified local segments to the unqualified local segment through a sparse segment graph. Specifically, a bipartite graph is constructed over the local segments, where qualified local segments $\mathcal{S}_i$ and unqualified local segments $\mathcal{S}_j$, with centroids $\boldsymbol{c}_i$ and $\boldsymbol{c}_j$, respectively, constitute two disjoint node sets. Edges are established only between these two types of nodes, with edge weights $w_{ij}$ defined by an exponential distance-decay function as follows:
\begin{equation}
w_{ij} =
\exp\left(-\|\boldsymbol{c}_i-\boldsymbol{c}_j\|_2^2/\tau_j\right),
\end{equation}
where $\tau_j$ denotes a scaling factor adaptively computed from the median distance between the local segment $\mathcal{S}_j$ and its selected neighboring segments.
The bipartite graph is then sparsified by retaining only $K$ edges with the highest weights for each unqualified local segment $\mathcal{S}_j$, with the indices of the corresponding local segments denoted by $\mathcal{N}_j$.
This sparsification removes distant connections, thereby reducing computational cost and preventing unreliable parameter propagation from spatially irrelevant regions.
The unknown affine transformation parameters $\hat{\boldsymbol{\theta}}_j$ of each unqualified local segment $\mathcal{S}_j$ are finally recovered by aggregating the parameters of its neighboring qualified local segments using the following expression:
\begin{equation}
\hat{\boldsymbol{\theta}}_j=
\frac{
\sum_{i\in\mathcal N_j}w_{ij}\hat{\boldsymbol{\theta}}_i
}
{
\sum_{i\in\mathcal N_j}w_{ij}
},
\quad j\notin\mathcal Q.
\end{equation}
In this way, affine transformation parameters estimated from qualified local segments can be propagated to regions without sufficient metric observations, enabling more reliable parameter estimation with support from neighboring segments.

\subsection{Explicit Depth Refinement via Geodesic Dynamic Programming}
\label{sect.geodesic}

The above scale recovery process enables the relative depth within each segment to be metricized via a shared affine transformation while preserving the original local geometric structure in the resulting metric depth. Nevertheless, such a region-wise depth metricization strategy also prevents the metric observations from being exactly preserved at corresponding pixels, possibly introducing local errors at or around these observations. In contrast, depth completion approaches~\cite{marigolddc, ognidc, omnidc} typically treat sparse metric observations as explicit anchors and infer metric depth around them, with relative depth serving as an auxiliary geometric prior to guide depth inference. Although such approaches can minimize local errors around the observations more directly, the fine-grained structures in the original relative depth maps cannot be fully preserved during depth completion. To combine the strengths of these two paradigms and achieve interpretable and generalizable depth metricization while preserving the fine-grained geometric structures in the original relative depth maps, this study proposes an explicit observation-guided pixel-wise depth refinement paradigm. This paradigm utilizes metric observations as anchors, progressively propagates metric depth information to neighboring pixels through local surface modeling, and iteratively fuses the propagated depth estimates with the current depth values.

Consider one propagation step from a neighboring pixel $\boldsymbol{q} \in \mathcal{N}_{\boldsymbol{p}}$ to a target pixel $\boldsymbol{p}$, where $\mathcal{N}_{\boldsymbol{p}}$ denotes the set of eight pixels adjacent to $\boldsymbol{p}$. When the local depth at $\boldsymbol{p}$ is estimated from $\boldsymbol{q}$ using a first-order surface approximation \cite{geonet, Zhao_2021_ICCV}, the resulting approximation error can be quantified by the following remainder term $R(\boldsymbol{p},\boldsymbol{q})$\footnote{Compared with the conventional first-order remainder $R(\boldsymbol{p},\boldsymbol{q})=z(\boldsymbol{q})-z(\boldsymbol{p})-\nabla{z}(\boldsymbol{p})^\top(\boldsymbol{q}-\boldsymbol{p})$, the remainder in~\eqref{eq:first-order-remainder} is antisymmetric, \iei, $R(\boldsymbol{p},\boldsymbol{q})+R(\boldsymbol{q},\boldsymbol{p})=0$, and therefore $|R(\boldsymbol{p},\boldsymbol{q})| = |R(\boldsymbol{q},\boldsymbol{p})|$. This antisymmetry eliminates orientation bias and isolates second-order variation. In particular, $R(\boldsymbol{p},\boldsymbol{q})=0$ for any affine depth field $z$.}:
\begin{equation}
\label{eq:first-order-remainder}
R(\boldsymbol{p},\boldsymbol{q})
\triangleq
z(\boldsymbol{q})-z(\boldsymbol{p})
-
\frac{1}{2}
\left(
\nabla{z}(\boldsymbol{p})+\nabla{z}(\boldsymbol{q})
\right)^\top
(\boldsymbol{q}-\boldsymbol{p}).
\end{equation}
Let $\mathcal{P}_{\boldsymbol{p}\to\boldsymbol{q}}$ denote the set of shortest paths connecting adjacent pixels $\boldsymbol{p}$ and $\boldsymbol{q}$, with each path consisting of only horizontal and vertical line segments. If $\boldsymbol{p}$ and $\boldsymbol{q}$ are horizontally or vertically adjacent, this set contains a single direct path. If they are diagonally adjacent, it contains two shortest paths, corresponding to the horizontal-then-vertical and vertical-then-horizontal connections.
As proven in our preliminary work~\cite{mtd_cvpr}, if the second partial derivatives $z_{uu}$ and $z_{vv}$ exist and are integrable, the magnitude of the remainder is bounded by the accumulated second-order depth variation along each path $L\in\mathcal{P}_{\boldsymbol{p}\to\boldsymbol{q}}$ as follows:
\begin{equation}
\bigl|R(\boldsymbol{p},\boldsymbol{q}) \bigr|
\le
\int_{L}\phi(u,v)\,ds, 
\label{eq:path-bound}
\end{equation}
where
\begin{equation}
\phi(u,v)
\triangleq
\sqrt{z_{uu}^2(u,v)+z_{vv}^2(u,v)}.
\label{eq:density}
\end{equation}
Accordingly, the tightest bound over these local paths is given by the following expression:
\begin{equation}
\label{eq:def-dphi}
g_\phi(\boldsymbol{p},\boldsymbol{q})
=
\min_{L\in\mathcal{P}_{\boldsymbol{p}\to\boldsymbol{q}}}
\int_{L}\phi(u,v)\,ds,
\end{equation}
which defines a discontinuity-aware geodesic cost for each local depth propagation step. The propagation route from a pixel with a valid metric observation to a target pixel can then be determined by minimizing the accumulated geodesic cost over all local propagation steps along the route.

In practice, finding the optimal depth propagation route can naturally be formulated as a dynamic programming problem, since the optimal route to a target pixel can be recursively determined from the optimal routes of its neighboring pixels.
Specifically, a propagation route is represented by a sequence of neighboring pixels $\Pi:\boldsymbol{p}^{(0)}\to\boldsymbol{p}^{(1)}\to\cdots\to\boldsymbol{p}^{(K)}$, where $\boldsymbol{p}^{(0)}$ is a pixel with a valid metric observation, $\boldsymbol{p}^{(K)}$ is the terminal pixel of the $K$-step route, and each transition $\boldsymbol{p}^{(k)}\to\boldsymbol{p}^{(k+1)}$ corresponds to one local depth propagation step. 
The continuous path integral in~\eqref{eq:def-dphi} can then be decomposed over the individual propagation steps and discretized as follows:
\begin{equation}
\label{eq:L_Pi}
\int_{L_\Pi}\phi\,ds
=
\sum_{k=0}^{K-1}
\int_{L^{(k)}}\phi\,ds
\approx
\sum_{k=0}^{K-1}
\underbrace{
\ell(\boldsymbol{p}^{(k)},\boldsymbol{p}^{(k+1)})
\phi(\boldsymbol{p}^{(k+1)})
}_{
W(\boldsymbol{p}^{(k)}\to\boldsymbol{p}^{(k+1)})
},
\end{equation}
where $L^{(k)}$ represents the path segment corresponding to the $k$-th propagation step $\boldsymbol{p}^{(k)}\to\boldsymbol{p}^{(k+1)}$, $L_\Pi$ denotes the complete path obtained by concatenating all such segments along the route $\Pi$, $\ell(\cdot,\cdot)$ represents a step-length function that computes the Euclidean distance between adjacent pixels, with each horizontal or vertical step assigned unit length, and $W(\boldsymbol{p}^{(k)}\to\boldsymbol{p}^{(k+1)})$ denotes the approximated geodesic cost of the corresponding propagation step.

At initialization, a depth map is constructed by assigning the metric observations to pixels with valid observations and treating these pixels as propagation seeds, while retaining the scale-recovered depth estimates at unobserved pixels\footnote{In practice, the local gradient and curvature are captured from the scale-recovered depth before the metric depth measurements are assigned, thereby preserving the original geometric structure and preventing the newly assigned depth measurements from giving rise to surface discontinuities.}. 
A cumulative cost map, denoted by $\boldsymbol{G}_{\phi}^{(t)}$, is simultaneously initialized to zero at these seeds and to a sufficiently large value at unobserved pixels.
The cumulative cost map $\boldsymbol{G}_{\phi}^{(t)}$ is then iteratively updated to record the minimum accumulated propagation cost to each pixel using the following expression:
\begin{equation}
\label{eq:dp_cost}
\boldsymbol{G}_\phi^{(t+1)}(\boldsymbol{p})
=
\min\left\{\boldsymbol{G}_\phi^{(t)}(\boldsymbol{p}), \min_{\boldsymbol{q}\in\mathcal{N}_{\boldsymbol{p}}}
\left[
\boldsymbol{G}_\phi^{(t)}(\boldsymbol{q})
+
W(\boldsymbol{q}\to\boldsymbol{p})
\right]\right\}.
\end{equation}
For each neighboring pixel $\boldsymbol{q}$, the term $\boldsymbol{G}_{\phi}^{(t)}(\boldsymbol{q})+W(\boldsymbol{q}\rightarrow\boldsymbol{p})$ represents the accumulated cost of reaching $\boldsymbol{p}$ through $\boldsymbol{q}$.
By comparing these costs over all neighboring pixels, the propagation route with minimum cost to $\boldsymbol{p}$ is identified.

Along the route with the minimum propagation cost, corresponding to the lowest accumulated geometric variation, the metric depth values from observed pixels are progressively propagated to more distant pixels while constraining the accumulation of approximation error.
For each local propagation step along the selected route, the preceding pixel $\boldsymbol{q}$ selected using~\eqref{eq:dp_cost} propagates its depth value to the target pixel $\boldsymbol{p}$ through the following surface modeling process:
\begin{equation}
\label{eq:local_second_order}
\hat{z}^{(t)}
\left(
\boldsymbol{p}
|
\boldsymbol{q},\Delta\boldsymbol{p}
\right)
=
z^{(t)}(\boldsymbol{q})
+
\nabla z^{(t)}(\boldsymbol{q})^{\top}\Delta\boldsymbol{p}
+
\frac{1}{2}
\Delta\boldsymbol{p}^{\top}
\boldsymbol{H}_{z}^{(t)}(\boldsymbol{q})
\Delta\boldsymbol{p},
\end{equation}
where $\Delta\boldsymbol{p}=\boldsymbol{p}-\boldsymbol{q}$, and $\boldsymbol{H}_z^{(t)}(\boldsymbol{q})$ denotes the Hessian of the current depth map at $\boldsymbol{q}$. The first-order term captures the local depth gradient, while the second-order term captures surface curvature. Accordingly, the preceding first-order analysis provides a practical criterion for evaluating propagation reliability, whereas the second-order local model improves the accuracy of depth prediction along the selected route. The resulting depth is then iteratively fused with the current depth, expressed as follows:
\begin{equation}
\label{eq:update}
z^{(t+1)}(\boldsymbol{p})
=
\left(1-\frac{1}{t+1}\right)
z^{(t)}(\boldsymbol{p}) +
\frac{1}{t+1}
\hat z^{(t)}
\left(
\boldsymbol{p}
|
\boldsymbol{q},\Delta\boldsymbol{p}
\right).
\end{equation}
Completing the above iterative metric depth refinement yields the final output of the optimization-based depth metricization process, which is further refined by the following data-driven metric depth refinement pipeline for minimizing residual depth errors.

\subsection{Data-Driven Depth Refinement via Learnable Multi-Path Propagation}
\label{sect.training_guided}
The aforementioned explicit depth refinement pipeline relies heavily on local surface modeling and predefined propagation rules. Nevertheless, it is often difficult to estimate local depth variations associated with thin structures, intricate geometric details, or irregular surfaces via local surface modeling, while the predefined propagation rules may make suboptimal propagation decisions in these challenging regions, inevitably introducing noticeable errors into the refined depth~\cite{spn1,nlspn,dyspn,bpnet}. By leveraging knowledge implicitly learned from large-scale training data, neural networks are particularly effective in estimating local depth variations in complex scenarios. Nonetheless, they can also be susceptible to overfitting the training data and are typically less interpretable~\cite{mtd_cvpr,ranftl2020midas,omnidc}.
To address these limitations, MTDv2 introduces data-driven network estimates built upon the explicit depth metricization process.

First, the explicitly metricized dense depth is introduced as a complementary input to the sparse or coarsely preprocessed dense depth maps commonly adopted by depth completion networks~\cite{ognidc,omnidc,cformer,lrru,bpnet,dmd3c}. Providing this complementary input relieves the network from the burden of directly recovering a complete metric depth map from sparse observations, thereby reducing its sensitivity to data distribution shifts. Furthermore, the network is designed to minimize residual errors arising from local surface modeling and to extend the original single-path depth propagation into a soft multi-path process, while the overall depth refinement pipeline largely follows the proposed explicit optimization-based paradigm to preserve its interpretability.

Specifically, the data-driven depth refinement adopts a two-stream feature extraction architecture, with one stream extracting textural features and the other extracting geometric features. Textural features are extracted from RGB images, while geometric features are extracted by jointly taking the output of the explicit metricization process, sparse observations, and a validity mask as input, thereby exploiting not only sparse observations used by previous approaches but also dense metric depth with fine-grained geometric structures. The textural and geometric features are then fused and fed into a geometry-guided depth propagation head, which predicts the residuals of surface modeling and geodesic cost computation, as well as adaptive propagation weights to further refine the explicitly metricized depth through local error minimization and iterative depth propagation analogous to the explicit optimization-based paradigm introduced in Sect.~\ref{sect.geodesic}.

The depth map for data-driven refinement is initialized with the output of the explicit metricization process as $z^{(0)}(\boldsymbol{p})=z_{\mathrm{opt}}(\boldsymbol{p})$.
For each pixel $\boldsymbol{p}$, the depth $z^{(0)}(\boldsymbol{p}|\boldsymbol{q},\Delta\boldsymbol p)$ propagated from its adjacent pixel $\boldsymbol{q}$ is first computed via the local surface modeling process expressed in~\eqref{eq:local_second_order}.
Subsequently, at the $t$-th iteration, the geometry-guided depth propagation head predicts a depth propagation residual $r_z^{(t)}(\boldsymbol{p},\boldsymbol{q})$ to iteratively refine the propagated depth using the following expression:
\begin{equation}
\label{eq:candidate_residual}
z^{(t+1)}\left(\boldsymbol{p}|\boldsymbol{q},\Delta\boldsymbol{p}\right)
=
z^{(t)}\left(\boldsymbol{p}|\boldsymbol{q},\Delta\boldsymbol{p}\right)
+
\rho_z r_z^{(t)}(\boldsymbol{p},\boldsymbol{q}),
\end{equation}
where $\rho_z$ is a scaling factor that controls the magnitude of the residual update.
This enables visual cues extracted from the original RGB image to be leveraged to minimize local errors arising from the finite-order surface modeling, thereby improving the representation of complex local depth variations that are difficult to model explicitly.
Similarly, since the discontinuity density defined in~\eqref{eq:density}, which is used to compute the geodesic cost for quantifying propagation reliability, is derived from local second-order derivatives, it may also be insufficient to characterize complex discontinuities. 
Therefore, the discontinuity density, denoted as $\hat{\phi}^{(0)}(\boldsymbol{p})$, is initialized using~\eqref{eq:density} and is iteratively refined using the residual $r_{\phi}^{(t)}(\boldsymbol{p})$ predicted by the geometry-guided depth propagation head as follows:
\begin{equation}
\label{eq:energy_residual}
\hat{\phi}^{(t+1)}(\boldsymbol{p})
=
\max\!\left(
\epsilon,\,
\hat{\phi}^{(t)}(\boldsymbol{p})
+
\rho_{\phi}r_{\phi}^{(t)}(\boldsymbol{p})
\right),
\end{equation}
where the scaling factor $\rho_{\phi}$ controls the magnitude of the residual update, while the hyperparameter $\epsilon$ guarantees that the updated density remains positive. As a result, the discontinuity information used to guide depth propagation can also be adaptively refined based on the knowledge learned from data.

Furthermore, the explicit depth refinement approach in Sect.~\ref{sect.geodesic} selects only one neighboring pixel for depth propagation at each step, thereby discarding other locally plausible predictions that may provide complementary information. To address this limitation, MTDv2 formulates a soft depth propagation process based on affinity scores $a^{(t)}(\boldsymbol{p}, \boldsymbol{q})$ computed from the refined discontinuity density $\hat{\phi}^{(t)}$ and a propagation logit $\gamma^{(t)}(\boldsymbol{p}, \boldsymbol{q})$ predicted by the geometry-guided depth propagation head as follows:
\begin{equation}
\label{eq:sda_affinity}
a^{(t)}(\boldsymbol{p}, \boldsymbol{q})
=
\frac{
\exp\!\left(
\gamma^{(t)}(\boldsymbol{p}, \boldsymbol{q})
-
\lambda_{\phi}\hat{\phi}^{(t)}(\boldsymbol{q})
\right)
}{
\sum_{\boldsymbol{r}\in\mathcal N_{\boldsymbol{p}}}
\exp\!\left(
\gamma^{(t)}(\boldsymbol{p}, \boldsymbol{r})
-
\lambda_{\phi}\hat{\phi}^{(t)}(\boldsymbol{r})
\right)
},
\end{equation}
where $a^{(t)}(\boldsymbol{p}, \boldsymbol{q})$ denotes the weight assigned to the depth propagated from $\boldsymbol{q}$ to $\boldsymbol{p}$, and $\lambda_{\phi}$ controls the contribution of the discontinuity density to the affinity score computation. The depth propagated from each adjacent pixel is then aggregated according to its corresponding affinity score and fused with the current depth as follows:
\begin{equation}
\label{eq:sda_guided_update}
\hat z^{(t+1)}(\boldsymbol{p})
= 
(1-\eta_t)z^{(t)}(\boldsymbol{p})
+
\eta_t
\sum_{\boldsymbol{q}\in\mathcal N_{\boldsymbol{p}}}
a^{(t)}(\boldsymbol{p}, \boldsymbol{q})
z^{(t)}
\left(
\boldsymbol{p}|\boldsymbol{q},\Delta\boldsymbol{p}
\right),
\end{equation}
where the hyperparameter $\eta_t$ balances the contributions of the current depth and the propagated depth. Repeated application of this propagation step yields an implicit multi-path depth propagation process that integrates depth estimates propagated from multiple locally plausible directions, thereby reducing sensitivity to potential erroneous predecessor selections.

Finally, the available metric observations are used to anchor the refined depth at each iteration, with an anchoring confidence $c(\boldsymbol{p})$ predicted by the geometry-guided depth propagation head as follows:
\begin{equation}
\label{eq:sda_update_anchor}
z^{(t+1)}(\boldsymbol{p})
=
\begin{cases}
(1-c(\boldsymbol{p}))
\hat z^{(t+1)}(\boldsymbol{p})
+
c(\boldsymbol{p})o(\boldsymbol{p}),
&
m(\boldsymbol{p})=1,
\\[0.5ex]
\hat z^{(t+1)}(\boldsymbol{p}),
&
m(\boldsymbol{p})=0.
\end{cases}
\end{equation}
which continuously reinforces the measured metric depth at observed locations while propagating these reliable constraints to guide depth refinement in surrounding regions during subsequent iterations.
The above data-driven metric depth refinement yields the final output of MTDv2, leveraging the strengths of deep learning to compensate for the limitations of optimization-based depth metricization while preserving the interpretability of the explicit formulation.

\subsection{Pseudo Label-Based Depth Propagation Head Training}
\label{sect.hybrid_training}

Conventional data-driven depth completion approaches~\cite{dyspn,bpnet,cformer,ognidc,omnidc} typically require large-scale accurate dense metric depth annotations for network training. Acquiring such annotations in real-world environments remains challenging because commonly used range sensing modalities either provide accurate yet sparse measurements, such as the point clouds acquired by LiDAR or radar sensors, or dense measurements affected by limited range and sensor-specific artifacts, as commonly encountered with RGB-D camera and time-of-flight sensors. Existing methods therefore generally rely either on relatively limited real-world datasets~\cite{nyu, kitti} or large-scale synthetic datasets~\cite{vkitti2, hypersim, tartanair}. Nevertheless, training on limited real-world data often restricts generalization across diverse scenes, whereas large-scale synthetic datasets, despite improving scene diversity, inevitably suffer from the domain gap between synthetic and real-world data~\cite{ranftl2020midas,depthanything,depthanythingv2}, thereby limiting model performance in real-world applications.

In MTDv2, dense depth maps with recovered metric scales are already provided by the explicit metricization process, eliminating the need for the network to learn globally accurate metric scales from training data and thereby enabling effective supervision using large-scale pseudo metric depth labels.
Although the pseudo metric depth labels may contain considerable local depth errors or exhibit global distribution shifts, the sparse observations sampled from these labels naturally inherit the same errors and shifts.
By applying the explicit depth refinement strategy to these sparse observations, the relationship between the explicitly metricized depth and the pseudo labels remains consistent with that encountered in real-world scenarios. Therefore, their discrepancy reflects the residual of the explicit refinement process and provides effective supervision for training the subsequent geometry-guided propagation head.

MTDv2 can therefore be trained using unlabeled real-world images with pseudo metric depth labels, without imposing stringent requirements on the reliability of these pseudo labels. Specifically, in addition to high-quality synthetic datasets including TartanAir~\cite{tartanair}, VKITTI2~\cite{vkitti2}, Hypersim~\cite{hypersim}, IRS~\cite{wang2021irs}, and BlendedMVS~\cite{yao2020blendedmvs}, diverse unlabeled real-world images from the BDD100K~\cite{yu2020bdd100k}, SA-1B~\cite{sam}, and MegaDepth~\cite{li2018megadepth} datasets are incorporated into the training set. Their dense pseudo metric depth labels are generated using the high-performing metric depth estimation model~\cite{moge2}. These dense pseudo labels are subsequently sampled according to different observation patterns to simulate various real-world sparse depth acquisition mechanisms, producing the corresponding sparse metric depth observations for training. Additionally, samples with insufficient valid pixels or limited depth variation are discarded because they provide insufficient geometric information for learning meaningful local depth refinement. This data filtering strategy is particularly important for images in the SA-1B dataset that predominantly contain planar regions, such as paintings, posters, and stamps. In such cases, the outputs of the explicit metricization process may already be close to the dense pseudo labels, leaving only a small residual discrepancy and thereby providing insufficient supervisory signals for training the propagation head.

\section{Experiments}
\label{sect.experiments}

\subsection{Implementation Details}
\label{sect.implementation}

Within the optimization-based depth metricization pipeline, Depth~Anything~V2 (DAV2)~\cite{depthanythingv2} with a ViT-L backbone is employed as the default relative depth estimation network. For the experiments in Sects.~\ref{sect.general_recovery} and~\ref{sect.efficient}, different depth estimation networks are adopted according to the specific evaluation objectives. For the explicit depth refinement process in Sect.~\ref{sect.geodesic}, spatial derivatives are computed using finite differences, and the maximum number of depth propagation steps is empirically set to 3. An ablation study on the number of depth propagation steps is presented in Sect.~\ref{sect.ablation}. 

Within the data-driven metric depth refinement pipeline, the scaling factors $\rho_z$, $\rho_\phi$, and $\lambda_\phi$, defined in~\eqref{eq:candidate_residual},~\eqref{eq:energy_residual}, and~\eqref{eq:sda_affinity}, are set to $0.1$, $0.02$, and $0.05$, respectively. The positive constant $\epsilon$ in~\eqref{eq:energy_residual} is set to $10^{-6}$, while the update factor $\eta_t$ in~\eqref{eq:sda_guided_update} is progressively decreased over the depth propagation steps. Model training is conducted on four NVIDIA RTX 4090 GPUs for 20 epochs with a batch size of 12. The Adam optimizer is employed with an initial learning rate of $10^{-3}$. A multi-step learning rate scheduler is adopted, with the learning rate decayed by a factor of 0.3 every 5 epochs. Input images are resized and cropped to $480 \times 640$ pixels, with random scaling, horizontal flipping, and color jittering applied for data augmentation. The loss functions used to train the depth propagation head follow those adopted in the prior work~\cite{ognidc}.

\subsection{Datasets}

As MTDv2 is proposed as a general depth metricization approach, this study evaluates its performance both as a depth completion model and as a general depth scale recovery algorithm. In the depth completion experiments, MTDv2 is additionally evaluated using correspondence matches as sparse observations for depth metricization, and its performance is compared with SoTA stereo matching networks under a zero-shot setting. Moreover, the recovered metric depth is further applied to downstream tasks that rely on depth perception, with visual odometry and novel view synthesis selected as two representative applications to evaluate its broader applicability. Overall, these evaluations involve 16 public datasets. The datasets and corresponding evaluation protocols are summarized as follows:

\begin{itemize}
\item The \textbf{nuScenes~\cite{nuscenes}} dataset comprises large-scale autonomous driving data collected in diverse urban environments using multiple cameras, LiDAR, and radar sensors. It is used for the evaluation of zero-shot depth completion performance. Specifically, this study uses the test split, which contains 150 scenes with 6,019 front-camera images. Sparse metric observations are obtained by projecting the single-frame LiDAR measurements onto the corresponding image plane. Ground-truth depth maps are generated by aggregating LiDAR measurements from multiple temporal frames, following the prior work~\cite{Singh_2023_CVPR}. Samples exhibiting noticeable projection artifacts are discarded from the evaluation.

\item The \textbf{DDAD~\cite{ddad}} dataset contains diverse urban driving sequences collected using six synchronized cameras and long-range LiDAR sensors. It is used for the evaluation of zero-shot depth completion performance. Following the prior work~\cite{marigolddc}, this study uses 3,950 front-facing images from the official validation split for performance evaluation. Approximately $20\%$ of the available LiDAR depth measurements are sampled as sparse observations, using the same outlier filtering strategy as that adopted in the official KITTI Depth Completion~\cite{uhrig2017sparsity} benchmark.

\item The \textbf{Make3D~\cite{make3d}} dataset contains outdoor scenes that differ from those in typical autonomous driving datasets, with depth measurements acquired using a laser scanner. Following the previous study~\cite{dcpidepth}, its official test split of 134 images is used for the evaluation of zero-shot depth completion performance.

\item The \textbf{DIODE~\cite{diode}} dataset contains diverse indoor and outdoor scenes, with RGB images and accurate long-range depth measurements acquired using a unified sensor system, enabling complementary evaluation across different scene types. It is used to evaluate the performance of both depth completion and depth scale recovery. For depth completion, 325 indoor and 446 outdoor images are used. For depth scale recovery, the same evaluation split is adopted, with the valid depth range set to 256~m.

\item The \textbf{ETH3D~\cite{eth3d}} dataset provides high-resolution RGB images captured by DSLR cameras and a synchronized multi-camera rig, together with accurate geometric ground truth acquired using a high-precision laser scanner. It is first used to evaluate the performance of depth completion with sparse metric observations obtained from either range sensors or correspondence matching. For the former setting, 454 images are used, following the prior work~\cite{omnidc}. For the latter setting, this study follows the evaluation protocol adopted in the prior work~\cite{wen2025foundationstereo} and uses 27 stereo image pairs.
This dataset is further used to evaluate depth scale recovery and novel view synthesis performance. For depth scale recovery, the same split as that used for the depth completion is adopted, with the valid depth range set to 65~m. For novel view synthesis, the dataset is divided into a training set containing 13 scenes and a test set containing 12 scenes collected in diverse indoor and outdoor environments with complex geometric structures. Following the prior work~\cite{3dgs}, 1/8 of the views in each scene are randomly selected as test views while the remaining 7/8 are used as training views.

\item The \textbf{ScanNet~\cite{scannet}} dataset, covering a wide range of indoor room layouts and scene structures, is used to evaluate both depth completion and depth scale recovery performance. Following the previous study~\cite{marigold}, a total of 800 samples are selected for performance evaluation after removing frames containing a large proportion of invalid ground-truth depth pixels. For depth scale recovery, the same evaluation split is adopted, with the valid depth range set to 10~m.

\item The \textbf{VOID~\cite{void}} dataset provides real-world indoor and outdoor RGB-D data collected using an Intel RealSense sensor for sparse-to-dense depth completion. Sparse depth maps are provided with densities of 150, 500, and 1,500 points per frame. Following the previous study~\cite{marigolddc}, models are evaluated on the official test set comprising 800 frames, using the provided sparse depth maps with 1,500 points per frame.

\item The \textbf{SUN-RGBD~\cite{sunrgbd}} dataset contains diverse indoor scenes captured using four different RGB-D cameras, providing substantial variations in scene layouts and sensor characteristics. It is used to evaluate zero-shot depth completion performance. Samples that are also included in the NYUv2 dataset are excluded from the official test split, resulting in 4,395 images for performance evaluation.

\item The \textbf{HAMMER~\cite{hammer}} dataset contains challenging scenes involving reflective and transparent objects captured using a multimodal sensing system. It is used to evaluate zero-shot depth completion performance. Since no universally adopted evaluation protocol exists for this dataset, this study randomly samples 400 images from the official test scenes using a fixed random seed. These images will be released together with the source code upon publication to enable reproducibility.

\item The \textbf{iBims-1~\cite{ibims}} dataset provides accurate indoor geometry and fine-grained depth structures, with ground-truth geometry obtained using a highly accurate laser scanner. It is used to evaluate zero-shot depth completion performance. This study uses the official evaluation split containing 100 images.

\item The \textbf{KITTI~\cite{kitti}} dataset consists of 56 real-world driving scenarios collected using a stereo camera, with depth measurements derived from the LiDAR point clouds. This dataset is used to evaluate MTDv2 as a general depth scale recovery algorithm. Following the previous study~\cite{depthanythingv2}, this study adopts the Eigen evaluation split comprising 652 images and restricts the valid depth range to 80~m.

\item The \textbf{NYUv2~\cite{nyu}} dataset contains diverse indoor scenes captured using a Microsoft Kinect RGB-D sensor and is used to evaluate depth scale recovery performance. Following the previous study~\cite{depthanythingv2}, the standard test split containing 654 images is adopted, with the valid depth range set to 10~m.

\item The \textbf{Middlebury~\cite{middlebury}} dataset contains high-resolution stereo image pairs with accurate ground-truth disparity annotations obtained using a structured-light scanner. It is used to evaluate the performance of MTDv2 with sparse metric depth observations obtained from correspondence matches. Following the zero-shot evaluation protocol adopted in the prior work~\cite{wen2025foundationstereo}, 15 stereo image pairs are used.

\item The \textbf{KITTI Stereo 2012~\cite{kitti}} and \textbf{KITTI Stereo 2015~\cite{kitti2015}} datasets provide stereo image pairs from outdoor driving scenes with ground-truth disparity annotations. They are used to evaluate the performance of MTDv2 with sparse metric depth observations obtained from correspondence matches. The corresponding evaluation sets, containing 194 and 200 stereo image pairs, respectively, are adopted.

\item The \textbf{KITTI Odometry~\cite{geiger2013vision}} dataset provides 22 driving sequences, among which Seqs.~00--10 are accompanied by ground-truth camera poses, enabling quantitative trajectory evaluation. It is used to evaluate the metric depth produced by MTDv2 in the visual odometry task. Following the previous study~\cite{droid}, the sequences with available ground-truth poses are employed, excluding Seqs.~01, 03, and 04 because their relatively short trajectories provide limited information for the evaluation of the proposed method~\cite{dimode}.

\end{itemize}

For datasets without predefined sparse metric observations or a dataset-specific sampling protocol described above, this study synthesizes sparse inputs for depth completion by randomly retaining $0.01$--$0.1\%$ of the valid depth measurements. The sampling density is varied in the sensitivity analyses in Sect.~\ref{sect.efficient} to assess the robustness of MTDv2 under different levels of observation sparsity. For the evaluation of depth scale recovery performance, $1.0$--$1.5\%$ of the ground-truth depth measurements in each sample are retained as sparse metric observations for MTD and MTDv2.

\begin{table*}[t]
\centering
\settablefont
\setlength{\tabcolsep}{10.5pt}
\caption{{Quantitative comparisons of SoTA depth completion methods. The best results are shown in bold font.}}
\begin{tabular}{lcccccccccc}
\toprule
\multirow{2}{*}[-0.75ex]{Method} & \multicolumn{2}{c}{nuScenes} & \multicolumn{2}{c}{DDAD} &
\multicolumn{2}{c}{Make3D} & \multicolumn{2}{c}{DIODE} & \multicolumn{2}{c}{ETH3D} \\
\cmidrule(lr){2-3}\cmidrule(lr){4-5}\cmidrule(lr){6-7}\cmidrule(lr){8-9}\cmidrule(lr){10-11}
& RMSE $\downarrow$ & MAE $\downarrow$ & RMSE $\downarrow$ & MAE $\downarrow$
& RMSE $\downarrow$ & MAE $\downarrow$ & RMSE $\downarrow$ & MAE $\downarrow$
& RMSE $\downarrow$ & MAE $\downarrow$ \\
\midrule
\midrule
CFormer~\cite{cformer} \tiny\clgray{(CVPR'23)}        & 15.672 & 10.528 & 9.606 & 3.328 & 10.749 & 5.035 & 2.297 & 0.976 & 2.796 & 1.940 \\
LRRU~\cite{lrru} \tiny\clgray{(ICCV'23)}              & 13.660 & 8.472 & 9.164 & 2.738 & 13.023 & 5.893 & 2.795 & 1.947 & 3.016 & 2.337 \\
BP-Net~\cite{bpnet}  \tiny\clgray{(CVPR'24)}          & 15.092 & 10.592 & 8.903 & 2.712 & 12.034 & 5.353 & 2.724 & 1.831 & 3.365 & 2.665 \\
DMD$^3$C~\cite{dmd3c} \tiny\clgray{(CVPR'25)}         & 5.556  & 3.112 & 7.766 & 2.498 & 12.019 & 5.575 & 2.262 & 1.127 & 0.935 & 0.285 \\
PromptDA~\cite{promptda}  \tiny\clgray{(CVPR'25)}     & 9.072  & 5.325 & 8.487 & 2.891 & 12.392 & 5.803 & 2.139 & 1.287 & 1.172 & 0.479 \\
Marigold-DC~\cite{marigolddc} \tiny\clgray{(ICCV'25)} & 4.924  & 2.595 & 6.449 & 2.364 & 8.926 & 4.932 & 1.987 & 0.887 & 0.706 & 0.245 \\
\textbf{MTD}~\cite{mtd_cvpr} \tiny\clgray{(CVPR'26)}                          & 4.387 & 2.177 & \textbf{5.252} & 1.834 & 8.581 & 4.776 & 1.736 & 0.761 & 0.662 & 0.177 \\
\textbf{MTDv2}                                 & \textbf{3.081} & \textbf{1.344} & 5.470 & \textbf{1.095} & \textbf{8.225} & \textbf{2.634} & \textbf{1.534} & \textbf{0.461} & \textbf{0.384} & \textbf{0.108} \\
\midrule
\midrule
\multirow{2}{*}[-0.75ex]{Method} & \multicolumn{2}{c}{ScanNet} & \multicolumn{2}{c}{VOID1500} &
\multicolumn{2}{c}{SUN-RGBD} & \multicolumn{2}{c}{HAMMER} & \multicolumn{2}{c}{IBims-1}\\
\cmidrule(lr){2-3}\cmidrule(lr){4-5}\cmidrule(lr){6-7}\cmidrule(lr){8-9}\cmidrule(lr){10-11}
& RMSE $\downarrow$ & MAE $\downarrow$ & RMSE $\downarrow$ & MAE $\downarrow$
& RMSE $\downarrow$ & MAE $\downarrow$ & RMSE $\downarrow$ & MAE $\downarrow$
& RMSE $\downarrow$ & MAE $\downarrow$ \\
\midrule
\midrule
CFormer~\cite{cformer} \tiny\clgray{(CVPR'23)}        & 0.223 & 0.118 & 0.726 & 0.261 & 0.442 & 0.218 & 0.101 & 0.055 & 0.177 & 0.040 \\
LRRU~\cite{lrru} \tiny\clgray{(ICCV'23)}              & 1.175 & 0.773 & 0.698 & 0.232 & 1.304 & 0.808 & 1.418 & 1.137 & 0.298 & 0.107 \\
BP-Net~\cite{bpnet}  \tiny\clgray{(CVPR'24)}          & 1.326 & 1.055 & 0.704 & 0.230 & 1.327 & 1.003 & 1.620 & 1.415 & 0.302 & 0.119 \\
DMD$^3$C~\cite{dmd3c} \tiny\clgray{(CVPR'25)}         & 0.152 & 0.070 & 0.676 & 0.225 & 0.423 & 0.106 & 0.098 & 0.047 & 0.286 & 0.083 \\
PromptDA~\cite{promptda}  \tiny\clgray{(CVPR'25)}     & 0.179 & 0.072 & 0.605 & 0.191 & 0.264 & 0.095 & 0.106 & 0.070 & 0.308 & 0.122 \\
Marigold-DC~\cite{marigolddc} \tiny\clgray{(ICCV'25)} & 0.145 & 0.059 & 0.505 & 0.151 & 0.238 & 0.067 & 0.054 & 0.037 & 0.176 & \textbf{0.038} \\
\textbf{MTD}~\cite{mtd_cvpr} \tiny\clgray{(CVPR'26)}                            & 0.129 & 0.049 & 0.366 & 0.138 & 0.220 & 0.050 & 0.093 & 0.034 & 0.190 & 0.072 \\
\textbf{MTDv2}                                 & \textbf{0.095} & \textbf{0.033} & \textbf{0.358} & \textbf{0.132} & \textbf{0.125} & \textbf{0.033} & \textbf{0.032} & \textbf{0.020} & \textbf{0.169} & 0.063 \\
\bottomrule
\end{tabular}\label{tab:main_results_dc}
\end{table*}

\begin{figure*}[!t]
	\centering
	\includegraphics[width=0.99\textwidth]{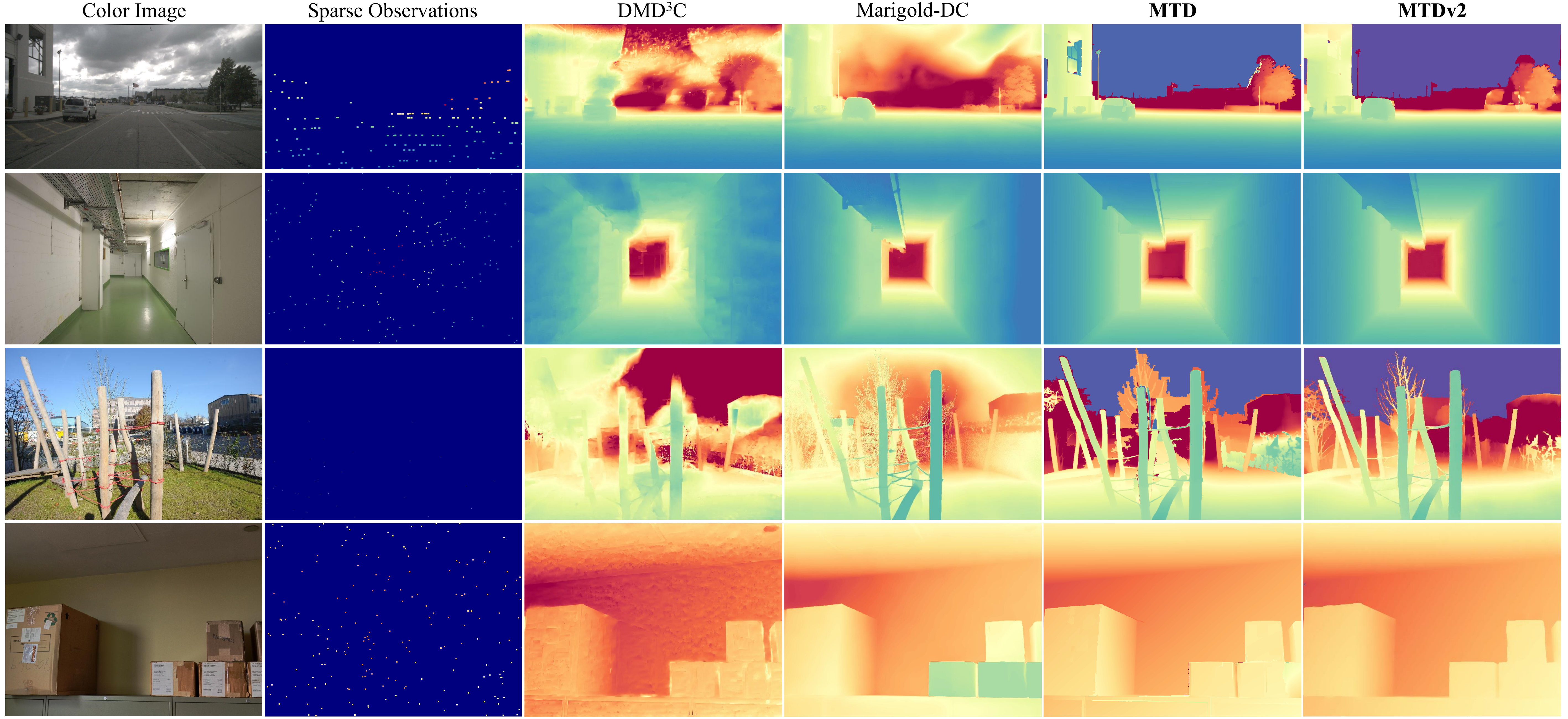}
	\caption{Qualitative comparisons with SoTA depth completion methods.}
\label{fig:main_results_dc}
\end{figure*}

\begin{table*}[t]
\begin{center}
\settablefont
\setlength{\tabcolsep}{7.5pt}
\caption{Quantitative comparison among the global scale recovery strategy, MTD, and the proposed MTDv2. The best results in each comparison group are shown in bold font.}
\begin{tabular}{clccccccccccc}
\toprule
\multirow{2}{*}[-0.75ex]{\makecell[c]{Model\\Paradigm}} &
\multirow{2}{*}[-0.75ex]{Method} &
\multirow{2}{*}[-0.75ex]{\makecell[c]{Configurations}} &
\multicolumn{2}{c}{KITTI}   & \multicolumn{2}{c}{NYUv2} &
\multicolumn{2}{c}{ETH3D}   & \multicolumn{2}{c}{ScanNet} &
\multicolumn{2}{c}{DIODE} \\
\cmidrule(lr){4-5}\cmidrule(lr){6-7}\cmidrule(lr){8-9}\cmidrule(lr){10-11}\cmidrule(lr){12-13}
& & & Abs Rel $\downarrow$ & $\delta_1$ $\uparrow$
  & Abs Rel $\downarrow$ & $\delta_1$ $\uparrow$
  & Abs Rel $\downarrow$ & $\delta_1$ $\uparrow$
  & Abs Rel $\downarrow$ & $\delta_1$ $\uparrow$
  & Abs Rel $\downarrow$ & $\delta_1$ $\uparrow$ \\
\midrule
\midrule
\multirow{15}{*}{\rotatebox{90}{\makecell[c]{\settablefont Discriminative Relative Depth \\ Estimation Network}}}
& MiDaS~\cite{birkl2023midasv3}
&              & 0.183 & 0.711 & 0.095 & 0.915 & 0.190 & 0.884 & 0.099 & 0.907 & 0.266 & 0.713 \\
& MiDaS~\cite{birkl2023midasv3}
& \gc{+MTD}   & \gc{0.069} & \gc{0.929} & \gc{0.048} & \gc{0.949} & \gc{0.055} & \gc{0.944} & \gc{0.015} & \gc{0.991} & \gc{0.113} & \gc{0.864} \\
& MiDaS~\cite{birkl2023midasv3}
& \gctr{+MTDv2} & \gctr{\textbf{0.053}} & \gctr{\textbf{0.938}} & \gctr{\textbf{0.041}} & \gctr{\textbf{0.956}} & \gctr{\textbf{0.047}} & \gctr{\textbf{0.952}} & \gctr{\textbf{0.012}} & \gctr{\textbf{0.993}} & \gctr{\textbf{0.102}} & \gctr{\textbf{0.871}} \\

& LeReS~\cite{leres}
&              & 0.149 & 0.784 & 0.090 & 0.916 & 0.171 & 0.777 & 0.091 & 0.917 & 0.271 & 0.766 \\
& LeReS~\cite{leres}
& \gc{+MTD}   & \gc{0.035} & \gc{0.974} & \gc{0.014} & \gc{0.994} & \gc{0.022} & \gc{0.984} & \gc{0.013} & \gc{0.994} & \gc{0.097} & \gc{0.885} \\
& LeReS~\cite{leres}
& \gctr{+MTDv2} & \gctr{\textbf{0.029}} & \gctr{\textbf{0.979}} & \gctr{\textbf{0.011}} & \gctr{\textbf{0.996}} & \gctr{\textbf{0.018}} & \gctr{\textbf{0.989}} & \gctr{\textbf{0.010}} & \gctr{\textbf{0.996}} & \gctr{\textbf{0.089}} & \gctr{\textbf{0.892}} \\

& DPT~\cite{dpt}
&              & 0.111 & 0.881 & 0.091 & 0.919 & 0.115 & 0.929 & 0.084 & 0.932 & 0.269 & 0.730 \\
& DPT~\cite{dpt}
& \gc{+MTD}   & \gc{0.032} & \gc{0.976} & \gc{0.016} & \gc{0.994} & \gc{0.019} & \gc{0.985} & \gc{0.014} & \gc{0.994} & \gc{0.096} & \gc{0.887} \\
& DPT~\cite{dpt}
& \gctr{+MTDv2} & \gctr{\textbf{0.026}} & \gctr{\textbf{0.980}} & \gctr{\textbf{0.013}} & \gctr{\textbf{0.996}} & \gctr{\textbf{0.015}} & \gctr{\textbf{0.989}} & \gctr{\textbf{0.011}} & \gctr{\textbf{0.996}} & \gctr{\textbf{0.088}} & \gctr{\textbf{0.890}} \\

& Depth Pro~\cite{depthpro}
&              & 0.077 & 0.949 & 0.044 & 0.975 & 0.060 & 0.965 & 0.042 & 0.980 & 0.321 & 0.752 \\
& Depth Pro~\cite{depthpro}
& \gc{+MTD}   & \gc{0.034} & \gc{0.975} & \gc{0.012} & \gc{0.995} & \gc{0.028} & \gc{0.974} & \gc{0.020} & \gc{0.986} & \gc{0.093} & \gc{0.889} \\
& Depth Pro~\cite{depthpro}
& \gctr{+MTDv2} & \gctr{\textbf{0.028}} & \gctr{\textbf{0.979}} & \gctr{\textbf{0.009}} & \gctr{\textbf{0.997}} & \gctr{\textbf{0.023}} & \gctr{\textbf{0.979}} & \gctr{\textbf{0.016}} & \gctr{\textbf{0.989}} & \gctr{\textbf{0.085}} & \gctr{\textbf{0.893}} \\

& Depth~Anything~V2~\cite{depthanythingv2}
&              & 0.080 & 0.946 & 0.043 & 0.980 & 0.062 & 0.980 & 0.043 & 0.981 & 0.260 & 0.759 \\
& Depth~Anything~V2~\cite{depthanythingv2}
& \gc{+MTD}   & \gc{0.022} & \gc{0.987} & \gc{0.013} & \gc{0.995} & \gc{0.017} & \gc{0.988} & \gc{0.016} & \gc{0.991} & \gc{0.093} & \gc{0.917} \\
& Depth~Anything~V2~\cite{depthanythingv2}
& \gctr{+MTDv2} & \gctr{\textbf{0.018}} & \gctr{\textbf{0.991}} & \gctr{\textbf{0.009}} & \gctr{\textbf{0.995}} & \gctr{\textbf{0.015}} & \gctr{\textbf{0.991}} & \gctr{\textbf{0.012}} & \gctr{\textbf{0.994}} & \gctr{\textbf{0.086}} & \gctr{\textbf{0.922}} \\

\midrule

\multirow{12}{*}{\rotatebox{90}{\makecell[c]{\settablefont Generative Relative Depth \\ Estimation Network}}}
& Marigold~\cite{marigold}        
&              & 0.099 & 0.916 & 0.055 & 0.964 & 0.065 & 0.960 & 0.064 & 0.951 & 0.308 & 0.773 \\
& Marigold~\cite{marigold}        
& \gc{+MTD}   & \gc{0.041} & \gc{0.966} & \gc{0.017} & \gc{0.991} & \gc{0.025} & \gc{0.978} & \gc{0.014} & \gc{0.991} & \gc{0.094} & \gc{0.896} \\
& Marigold~\cite{marigold}        
& \gctr{+MTDv2} & \gctr{\textbf{0.034}} & \gctr{\textbf{0.971}} & \gctr{\textbf{0.014}} & \gctr{\textbf{0.993}} & \gctr{\textbf{0.021}} & \gctr{\textbf{0.983}} & \gctr{\textbf{0.011}} & \gctr{\textbf{0.993}} & \gctr{\textbf{0.087}} & \gctr{\textbf{0.901}} \\

& GeoWizard~\cite{geowizard}       
&              & 0.097 & 0.921 & 0.052 & 0.966 & 0.064 & 0.961 & 0.061 & 0.953 & 0.297 & 0.792 \\
& GeoWizard~\cite{geowizard}       
& \gc{+MTD}   & \gc{0.047} & \gc{0.962} & \gc{0.014} & \gc{0.994} & \gc{0.021} & \gc{0.984} & \gc{0.013} & \gc{0.994} & \gc{0.097} & \gc{0.885} \\
& GeoWizard~\cite{geowizard}       
& \gctr{+MTDv2} & \gctr{\textbf{0.040}} & \gctr{\textbf{0.967}} & \gctr{\textbf{0.011}} & \gctr{\textbf{0.996}} & \gctr{\textbf{0.017}} & \gctr{\textbf{0.989}} & \gctr{\textbf{0.010}} & \gctr{\textbf{0.996}} & \gctr{\textbf{0.089}} & \gctr{\textbf{0.890}} \\

& Lotus~\cite{lotus}            
&              & 0.093 & 0.928 & 0.053 & 0.967 & 0.068 & 0.953 & 0.060 & 0.963 & 0.228 & 0.738 \\
& Lotus~\cite{lotus}            
& \gc{+MTD}   & \gc{0.032} & \gc{0.975} & \gc{0.015} & \gc{0.993} & \gc{0.022} & \gc{0.985} & \gc{0.015} & \gc{0.992} & \gc{0.108} & \gc{0.868} \\
& Lotus~\cite{lotus}            
& \gctr{+MTDv2} & \gctr{\textbf{0.026}} & \gctr{\textbf{0.980}} & \gctr{\textbf{0.012}} & \gctr{\textbf{0.995}} & \gctr{\textbf{0.018}} & \gctr{\textbf{0.989}} & \gctr{\textbf{0.012}} & \gctr{\textbf{0.994}} & \gctr{\textbf{0.097}} & \gctr{\textbf{0.873}} \\

& DepthMaster~\cite{depthmaster}     
&              & 0.082 & 0.937 & 0.050 & 0.972 & 0.053 & 0.974 & 0.055 & 0.967 & 0.215 & 0.776 \\
& DepthMaster~\cite{depthmaster}     
& \gc{+MTD}   & \gc{0.043} & \gc{0.965} & \gc{0.016} & \gc{0.993} & \gc{0.021} & \gc{0.984} & \gc{0.016} & \gc{0.992} & \gc{0.095} & \gc{0.888} \\
& DepthMaster~\cite{depthmaster}     
& \gctr{+MTDv2} & \gctr{\textbf{0.036}} & \gctr{\textbf{0.970}} & \gctr{\textbf{0.013}} & \gctr{\textbf{0.995}} & \gctr{\textbf{0.017}} & \gctr{\textbf{0.989}} & \gctr{\textbf{0.013}} & \gctr{\textbf{0.994}} & \gctr{\textbf{0.086}} & \gctr{\textbf{0.893}} \\

\midrule

\multirow{6}{*}{\rotatebox{90}{\makecell[c]{\settablefont Metric Depth \\ Estimation Network}}}

& Metric3Dv2~\cite{hu2024metric3dv2}      
&              & 0.051 & 0.976 & 0.067 & 0.973 & 0.137 & 0.825 & 0.047 & 0.990 & 0.246 & 0.823 \\
& Metric3Dv2~\cite{hu2024metric3dv2}      
& \gc{+MTD}   & \gc{0.027} & \gc{0.980} & \gc{0.014} & \gc{0.994} & \gc{0.019} & \gc{0.986} & \gc{0.015} & \gc{0.992} & \gc{0.099} & \gc{0.881} \\
& Metric3Dv2~\cite{hu2024metric3dv2}      
& \gctr{+MTDv2} & \gctr{\textbf{0.022}} & \gctr{\textbf{0.984}} & \gctr{\textbf{0.011}} & \gctr{\textbf{0.996}} & \gctr{\textbf{0.015}} & \gctr{\textbf{0.990}} & \gctr{\textbf{0.012}} & \gctr{\textbf{0.994}} & \gctr{\textbf{0.091}} & \gctr{\textbf{0.887}} \\

& UniDepthV2~\cite{unidepth2}      
&              & 0.076 & 0.952 & 0.062 & 0.970 & 0.150 & 0.865 & 0.058 & 0.975 & 0.251 & 0.795 \\
& UniDepthV2~\cite{unidepth2}      
& \gc{+MTD}   & \gc{0.032} & \gc{0.977} & \gc{0.017} & \gc{0.993} & \gc{0.023} & \gc{0.980} & \gc{0.014} & \gc{0.993} & \gc{0.096} & \gc{0.887} \\
& UniDepthV2~\cite{unidepth2}      
& \gctr{+MTDv2} & \gctr{\textbf{0.026}} & \gctr{\textbf{0.982}} & \gctr{\textbf{0.014}} & \gctr{\textbf{0.995}} & \gctr{\textbf{0.019}} & \gctr{\textbf{0.985}} & \gctr{\textbf{0.011}} & \gctr{\textbf{0.995}} & \gctr{\textbf{0.088}} & \gctr{\textbf{0.892}} \\

\bottomrule
\end{tabular}
\label{tab:main_results_mde}
\end{center}
\end{table*}

\subsection{Evaluation Metrics}

Following previous studies~\cite{ognidc, marigolddc, omnidc}, root mean squared error (RMSE) and 
mean absolute error (MAE) are adopted to quantify MTDv2's performance on the depth completion task.
To compare MTDv2 with SoTA stereo matching approaches when sparse metric observations are obtained from correspondence matches, BP-1 and BP-2, which denote the percentages of valid pixels with absolute disparity errors greater than 1 and 2 pixels, respectively, are reported, and D1, which denotes the percentage of valid pixels whose disparity error exceeds both 3 pixels and 5\% of the corresponding disparity ground truth, is reported, following the prior study~\cite{wen2025foundationstereo}. When evaluating MTDv2 as a general depth scale recovery algorithm, commonly used metrics including absolute relative error (Abs Rel) and threshold-based accuracy $\delta_1$ are employed, following the previous study~\cite{depthanything}.

For the evaluation of downstream tasks, novel view synthesis quality is quantified using peak signal-to-noise ratio (PSNR), structural similarity (SSIM), and learned perceptual image patch similarity (LPIPS), following the prior study~\cite{3dgs}. Since the primary goal of evaluating this downstream task is to examine the benefit of leveraging recovered metric depth, RMSE between the rendered depth and ground-truth metric depth is additionally reported on the test views. In addition, visual odometry performance is quantified using the absolute trajectory error (ATE) in meters, following the previous work~\cite{droid}.

\subsection{Depth Completion Results}
\label{sect.depth_completion}

The proposed MTDv2 is first compared with SoTA depth completion methods~\cite{cformer, bpnet, lrru, dmd3c, marigolddc} to evaluate its zero-shot depth completion performance. Specifically, CFormer~\cite{cformer}, BP-Net~\cite{bpnet}, and LRRU~\cite{lrru} are evaluated using their official weights for the KITTI \cite{kitti} and NYUv2 \cite{nyu} datasets in outdoor and indoor scenarios, respectively. For DMD$^3$C~\cite{dmd3c}, this study uses its officially released weights for the model trained on large-scale datasets. Notably, the training data of DMD$^3$C include the nuScenes dataset. Since the official weights of PromptDA~\cite{promptda} are obtained on specific datasets and exhibit limited generalizability to unseen scenarios, PromptDA is retrained using the same training setting as MTDv2 for a fair comparison. Marigold-DC~\cite{marigolddc} is evaluated using its official weights, with the officially released efficient denoising configuration adopted for all experiments. Both MTD~\cite{mtd_cvpr} and MTDv2 are included for comparison.

The quantitative results in Table~\ref{tab:main_results_dc} show that both MTD and MTDv2 achieve SoTA performance among the compared depth completion approaches. In particular, MTDv2 reduces RMSE by $4.0\%$--$47.5\%$ compared with the previous best results, demonstrating the effectiveness of the proposed interpretable depth metricization paradigm for zero-shot depth completion. Compared with MTD, MTDv2 further reduces depth completion errors by a substantial margin across most datasets, highlighting the benefit of combining explicit depth metricization with data-driven metric depth refinement to exploit their complementary strengths. On the DDAD dataset, although MTDv2 shows a slight increase in RMSE, suggesting some larger depth errors may remain, it substantially reduces MAE, indicating improved overall depth accuracy. Moreover, previous approaches exhibit noticeable performance degradation on challenging datasets with complex scene structures, such as ScanNet, SUN-RGBD, and HAMMER, as well as on datasets with highly sparse metric observations, such as nuScenes. In contrast, MTDv2 maintains consistently strong performance under these challenging conditions, demonstrating superior robustness and generalizability. Furthermore, as illustrated in Fig.~\ref{fig:main_results_dc}, MTDv2 recovers more accurate metric depth with sharper object boundaries than previous approaches, particularly in regions with sparse observations, further demonstrating the effectiveness of the proposed paradigm.

In addition to acquiring sparse metric depth using range sensors, multi-view images with known relative camera poses can also provide accurate metric depth information. For instance, stereo image pairs with known baselines enable metric depth recovery through stereo matching. Nevertheless, acquiring large-scale ground-truth disparity annotations in real-world scenarios is labor-intensive~\cite{watson2020learning}, making the training and adaptation of stereo matching models to diverse real-world environments expensive. Conventional stereo matching also relies on accurate extrinsic calibration and stereo image rectification to constrain correspondences to the same epipolar line. In practice, calibration and rectification errors may violate this constraint and degrade dense stereo matching performance. Even with a well-trained model and accurate extrinsic calibration, dense correspondence matching can remain unreliable in challenging regions, such as areas with ambiguous textures and occlusions~\cite {igev_pp}. MTDv2 provides an alternative and more robust solution. Instead of directly estimating dense disparity maps, it first employs a general correspondence matching algorithm~\cite{loftr} to obtain sparse yet highly reliable correspondences. These correspondences are then converted into sparse metric depth observations, from which MTDv2 recovers dense metric depth. The resulting dense metric depth is subsequently converted back into disparity maps, enabling direct comparison with SoTA stereo matching networks~\cite{crestereo, dsmnet, maskcfnet, hvtraft, raft, selective_igev, igev, former_raft_dam, igev_pp, nmrf, wen2025foundationstereo}. All compared stereo matching networks are trained only on the Scene Flow~\cite{sceneflow} dataset.

The quantitative results in Table~\ref{tab:main_results_stereo} demonstrate that, without learning highly discriminative features specifically for stereo matching, MTDv2 still significantly outperforms most existing stereo matching approaches across diverse indoor and outdoor scenarios. Even compared with the representative vision foundation model-based stereo matching approach~\cite{wen2025foundationstereo}, which achieves strong performance in zero-shot disparity estimation, MTDv2 reduces the disparity error by 3.6\%, 9.4\%, and 16.3\% on the Middlebury, KITTI Stereo 2012, and KITTI Stereo 2015 datasets, respectively. MTDv2 also consistently outperforms MTD~\cite{mtd_cvpr} in this experiment. These results collectively demonstrate the effectiveness of this distinctive paradigm for recovering high-quality dense disparity maps. By using sparse yet reliable matches as metric observations, MTDv2 can effectively recover metric depth from generalizable relative depth estimates while reducing errors caused by unreliable correspondence matching, particularly in challenging regions.

\subsection{Depth Scale Recovery Results}
\label{sect.general_recovery}

Affine-invariant relative depth estimates cannot be directly used for depth perception tasks that require metric scale~\cite{unidepth2, hu2024metric3dv2}. They are typically aligned with ground-truth depth annotations through a global affine transformation using all valid metric observations~\cite{ranftl2020midas, depthanything}. From this perspective, MTDv2 can be treated as an improved scale recovery approach that enables spatially varying metricization using sparse metric observations and is therefore evaluated against the conventional global scale recovery strategy.
Meanwhile, despite significant progress in metric depth estimation approaches~\cite{unidepth2, hu2024metric3dv2}, maintaining accurate metric scale across highly diverse scenes remains challenging~\cite{mtd_cvpr}. Therefore, MTDv2 is further evaluated for its ability to refine metric depth estimates and reduce local depth errors by leveraging sparse metric observations.

As presented in Table~\ref{tab:main_results_mde}, 
the evaluation of the scale recovery performance involves three mainstream paradigms: discriminative relative depth estimation networks~\cite{birkl2023midasv3, leres, dpt, depthpro, depthanythingv2}, generative relative depth estimation networks~\cite{marigold, geowizard, lotus, depthmaster}, and metric depth estimation networks~\cite{unidepth2, hu2024metric3dv2}.
The results show that MTD~\cite{mtd_cvpr} can already effectively recover metric scale from diverse source depth estimates, while MTDv2 further achieves substantial improvements over MTD, demonstrating the effectiveness of extending the explicit depth metricization approach with a data-driven depth refinement pipeline. Notably, even earlier methods~\cite{birkl2023midasv3, leres}, which provide less accurate relative depth estimates than more recent approaches, can still be substantially improved to produce reliable metric depth, highlighting the robustness and broad compatibility of the proposed framework.

\subsection{Downstream Applications Results}

This subsection evaluates the effectiveness and applicability of the metric depth recovered by MTDv2 for downstream 3D perception tasks. Two representative applications, NVS and VO, are selected for evaluation. 

\subsubsection{Novel View Synthesis Results}

Novel view synthesis relies on accurate depth estimation to reconstruct consistent 3D scene geometry from multi-view images. To this end, the metric depth produced by MTDv2 and by previous depth completion methods \cite{bhat2023zoedepth, g2_monodepth, omnidc} is separately incorporated into the classical 3D Gaussian Splatting (3DGS)~\cite{3dgs} framework for performance comparison.

\begin{figure*}[!t]
	\centering
	\includegraphics[width=0.99\textwidth]{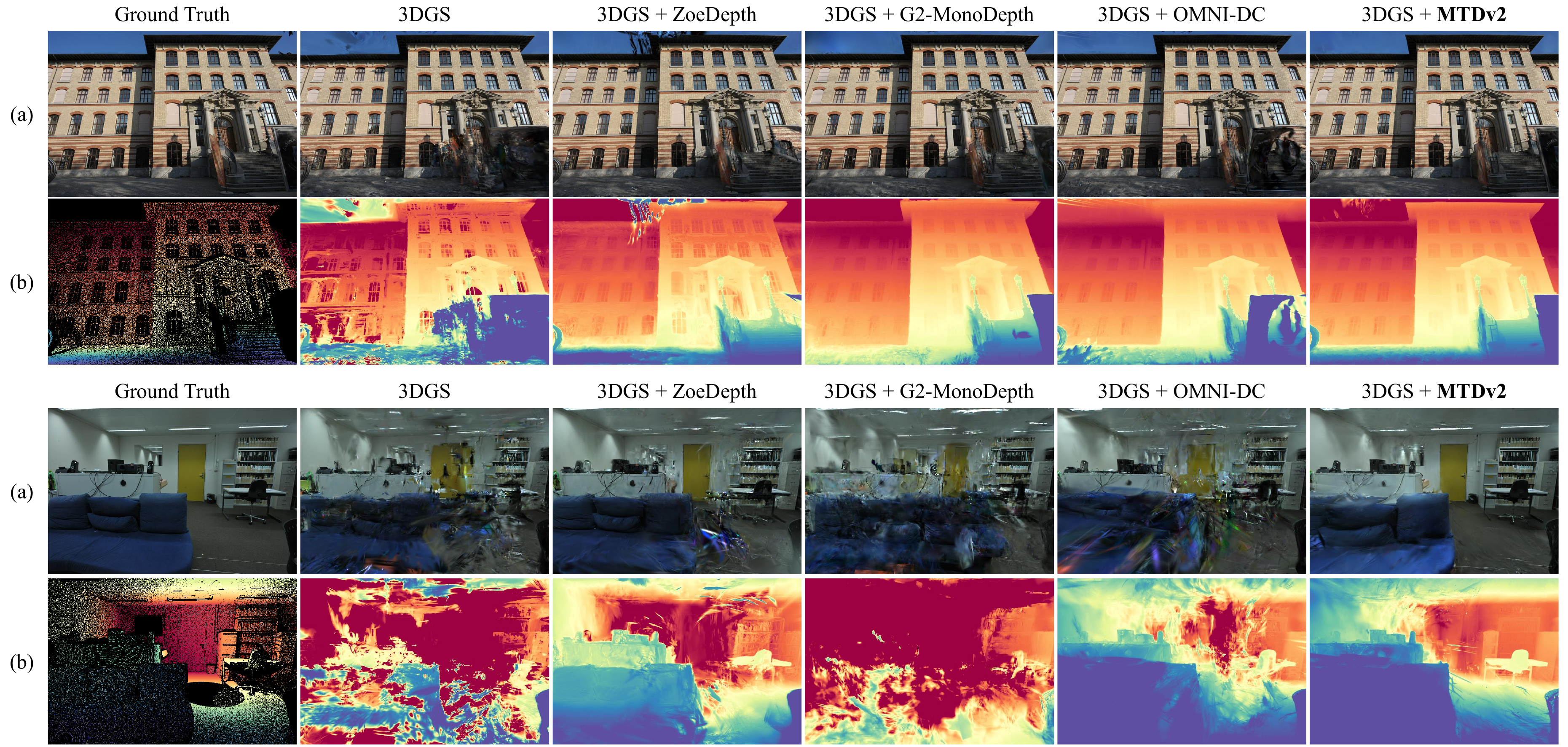}
	\caption{Qualitative comparisons of (a) rendered RGB images and (b) depth maps on test views obtained by the original 3DGS framework and by that incorporating different depth perception approaches.}
\label{fig:results_3dgs}
\end{figure*}

As presented in Table~\ref{tab:results_3dgs}, incorporating the metric depth produced by MTDv2 into the 3DGS pipeline achieves the best performance across all evaluation metrics, while substantially reducing the depth reconstruction error from 3.857 to 0.812 compared with the original 3DGS framework. These results demonstrate that accurate metric depth is critical for alleviating geometric ambiguity and improving the scene representation quality of 3DGS, while highlighting the advantage of MTDv2 over previous approaches in this downstream task.
This advantage is further evident from the qualitative results in Fig.~\ref{fig:results_3dgs}. Compared with the original 3DGS framework, incorporating MTDv2 into 3DGS produces rendered novel-view images and depth maps that are substantially closer to the ground truth. In contrast, incorporating previous depth completion methods into 3DGS leads to noticeable artifacts in both the synthesized novel-view images and the corresponding depth maps. These results indicate that the metric depth recovered by MTDv2 provides more reliable geometric information for downstream novel view synthesis, demonstrating its practical benefits beyond improving depth estimation accuracy alone.

\begin{table}[!t]
\centering
\settablefont
\setlength{\tabcolsep}{8.0pt}
\caption{Quantitative comparison with SoTA stereo matching networks for zero-shot disparity estimation. The best results are shown in bold font.}
\begin{tabular}{lcccc}
\toprule
\multirow{2}{*}{Method}
        & Middlebury & ETH3D & KITTI-12 & KITTI-15 \\
        & BP-2 $\downarrow$      & BP-1 $\downarrow$ & D1 $\downarrow$      & D1 $\downarrow$      \\
\midrule
\midrule
CREStereo++~\cite{crestereo}        & 14.8 & 4.4 & 4.7 & 5.2 \\
DSMNet~\cite{dsmnet}             & 13.8 & 6.2 & 6.2 & 6.5 \\
Mask-CFNet~\cite{maskcfnet}         & 13.7 & 5.7 & 4.8 & 5.8 \\
HVT-RAFT~\cite{hvtraft}           & 10.4 & 3.0 & 3.7 & 5.2 \\
RAFT-Stereo~\cite{raft}        & 12.6 & 3.3 & 4.7 & 5.5 \\
Selective-IGEV~\cite{selective_igev}     &  9.2 & 5.7 & 4.5 & 5.6 \\
IGEV~\cite{igev}               &  8.8 & 4.0 & 5.2 & 5.7 \\
Former-RAFT-DAM~\cite{former_raft_dam}    &  8.1 & 3.3 & 3.9 & 5.1 \\
IGEV++~\cite{igev_pp}             &  7.8 & 4.1 & 5.1 & 5.9 \\
NMRF~\cite{nmrf}               &  7.5 & 3.8 & 4.2 & 5.1 \\
FoundationStereo~\cite{wen2025foundationstereo}               &  5.5 & \textbf{1.8} & 3.2 & 4.9 \\
MTD~\cite{mtd_cvpr}              &  {6.1} & {2.8} & {3.6} & {5.0}\\
\textbf{MTDv2}               &  \textbf{5.3} & 2.2 & \textbf{2.9} & \textbf{4.1}\\
\bottomrule
\end{tabular}\label{tab:main_results_stereo}
\end{table}

\begin{table}[!t]
\centering
\settablefont
\setlength{\tabcolsep}{10.5pt}
\caption{Quantitative comparison of novel view synthesis methods on the ETH3D dataset. The best results are shown in bold font.}
\label{tab:eth3d_3dgs}
\begin{tabular}{lcccc}
\toprule
Method & PSNR$\uparrow$ & SSIM$\uparrow$ & LPIPS$\downarrow$ & RMSE $\downarrow$ \\
\midrule
\midrule
3DGS~\cite{3dgs}             & 15.64 & 0.557 & 0.418 & 3.857 \\
3DGS + ZoeDepth   & 18.96 & 0.573 & 0.324 & 2.163 \\
3DGS + G2-MonoDepth      & 19.36 & 0.641 & 0.273 & 1.904 \\
3DGS + OMNI-DC    & 20.38 & 0.660 & 0.229 & 0.838 \\
3DGS + \textbf{MTDv2}       & \textbf{20.42} & \textbf{0.689} & \textbf{0.213} & \textbf{0.812} \\
\bottomrule
\end{tabular}\label{tab:results_3dgs}
\end{table}

\begin{table}[!t]
\centering
\settablefont
\setlength{\tabcolsep}{8.5pt}
\caption{Quantitative comparison of visual odometry methods on the KITTI Odometry dataset. The best results are shown in bold font.}
\begin{tabular}{lcccc}
\toprule
\multirow{2}{*}{Method} & Seq 00 & Seq 02 & Seq 05 & Seq 06 \\
& ATE $\downarrow$      & ATE $\downarrow$ & ATE $\downarrow$      & ATE $\downarrow$      \\
\midrule
\midrule
DROID-SLAM (RGB)~\cite{droid}    &66.562 &84.828 &41.973 &72.402 \\
DROID-SLAM (RGB-D)          &63.295 &77.610 &31.641 &104.552 \\
DROID-SLAM (RGB-D) + \textbf{MTDv2}  &\textbf{24.985} &\textbf{26.050} &\textbf{12.513} &\textbf{13.504} \\
\midrule
\midrule
\multirow{2}{*}{Method} & Seq 07 & Seq 08 & Seq 09 & Seq 10 \\
& ATE $\downarrow$      & ATE $\downarrow$ & ATE $\downarrow$      & ATE $\downarrow$      \\
\midrule
\midrule
DROID-SLAM (RGB)~\cite{droid}    &13.785 &64.552 &60.414 &\textbf{11.394} \\
DROID-SLAM (RGB-D)          &9.867  &60.201 &42.493 &46.627 \\
DROID-SLAM (RGB-D) + \textbf{MTDv2}  &\textbf{6.138} &\textbf{20.965} &\textbf{24.832} &11.763 \\
\bottomrule
\end{tabular}\label{tab:results_vo}
\end{table}

\begin{figure*}[!t]
	\centering
	\includegraphics[width=0.99\textwidth]{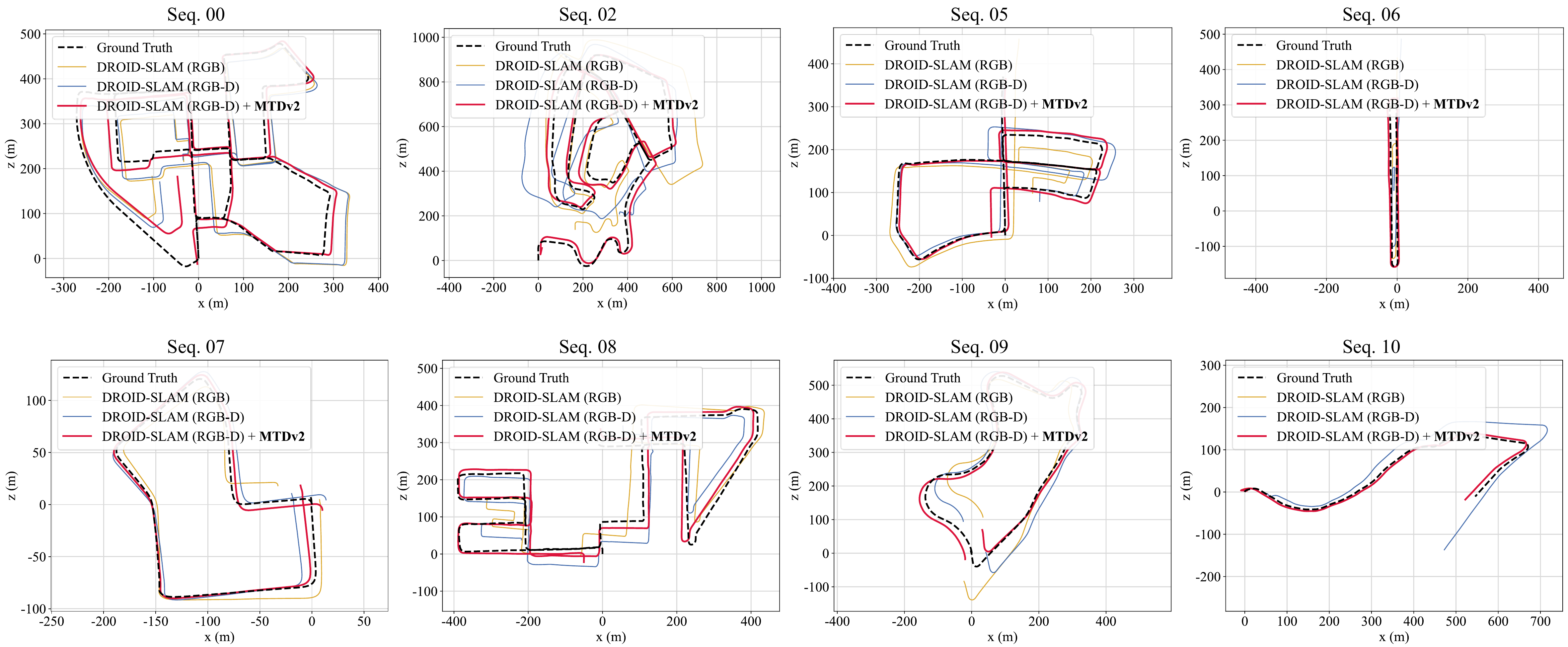}
	\caption{Qualitative comparisons of the visual odometry results achieved using RGB images only, RGB images with the original LiDAR measurements, and RGB images with dense metric depth recovered from the original sparse LiDAR measurements.
    }
\label{fig:results_vo}
\end{figure*}

\subsubsection{Visual Odometry Results}

Visual odometry requires metric depth information to estimate camera motion with metric scale and resolve scale ambiguity. This study adopts DROID-SLAM~\cite{droid} as the baseline framework to evaluate the applicability of the proposed MTDv2 to visual odometry. DROID-SLAM supports either RGB-only input or RGB-D input, where depth information can be obtained from LiDAR measurements. These sparse LiDAR measurements are further leveraged by MTDv2 to recover dense metric depth maps, which are then used as an alternative depth input to DROID-SLAM. 

As demonstrated in Table~\ref{tab:results_vo}, compared with the RGB-only configuration and the RGB-D configuration using the original LiDAR measurements, incorporating the metric depth recovered by MTDv2 reduces the average ATE by $66.2\%$ and $67.7\%$, respectively. 
These results demonstrate the effectiveness of recovering dense metric depth from the original sparse LiDAR measurements for visual odometry.
This improvement is further evident from the qualitative trajectory comparisons in Fig.~\ref{fig:results_vo}. The results show that incorporating the metric depth recovered by MTDv2 into DROID-SLAM produces trajectories substantially closer to the ground truth than those obtained using either baseline configuration, particularly for long sequences where accumulated drift becomes more pronounced.

\subsection{Ablation Studies}
\label{sect.ablation}

The ablation studies presented in Table~\ref{tab:ablation} validate the effectiveness of the four main components of MTDv2 by progressively incorporating them into the framework. The results of the configurations obtained after incorporating each component are reported in (a0)--(d0), respectively. Alternative settings and internal variants of each newly introduced component are further evaluated in the subsequent rows following the corresponding default configuration.

\begin{table}[t]
\centering
\settablefont
\setlength{\tabcolsep}{5.0pt}
\caption{Ablation studies on the KITTI and VOID datasets to validate the effectiveness of the proposed technical contributions in outdoor and indoor scenarios, respectively.}
\label{tab:ablation}
\begin{tabular}{llcccc}
\toprule
\multirow{2}{*}[-0.75ex]{} & \multirow{2}{*}[-0.75ex]{Configurations} &
\multicolumn{2}{c}{KITTI} & \multicolumn{2}{c}{VOID} \\
\cmidrule(lr){3-4}\cmidrule(lr){5-6}
 & & RMSE$\downarrow$ & MAE$\downarrow$
   & RMSE$\downarrow$ & MAE$\downarrow$ \\
\hline
\multicolumn{6}{c}{+ Segment-Wise Affine Transformation} \\
\hline
(a0)   & w/ LSF (inverse depth) & \textbf{6.782}  & \textbf{1.794} & \textbf{0.614} & \textbf{0.238} \\
(a1)   & w/ LSF           & 7.013  & 1.802 & 0.791 & 0.307 \\
(a2)   & w/ Median                  & 10.891 & 2.169 & 0.898 & 0.358 \\
(a3)   & w/ Quantile matching       & 9.178  & 2.114 & 0.821 & 0.346 \\
(a4)   & w/ Moment matching         & 8.756  & 1.983 & 0.815 & 0.329 \\

\hline
\multicolumn{6}{c}{+ Sparse Graph Optimization} \\
\hline
(b0)   & w/ Graph optimization  & \textbf{2.201} & \textbf{0.597} & \textbf{0.452} & \textbf{0.148} \\
(b1)   & w/ Global parameter        & 2.521 & 0.687 & 0.554 & 0.169 \\

\hline
\multicolumn{6}{c}{+ Explicit Depth Refinement} \\
\hline
(c0)   & MTD                             & 2.028 & 0.532 & \textbf{0.366} & \textbf{0.138} \\
(c1)   & w/ Laplacian-Based Cost               & 2.618 & 0.661 & 0.482 & 0.158 \\
(c2)   & w/ B-spline Interpolation             & 2.112 & 0.578 & 0.442 & 0.147 \\
(c3)   & Max. propagation steps = 1           & 2.049 & 0.551 & 0.429 & 0.145 \\
(c4)   & Max. propagation steps = 5                     & \textbf{1.913} & \textbf{0.498} & 0.391 & 0.141 \\

\hline
\multicolumn{6}{c}{+ Data-Driven Depth Refinement} \\
\hline
(d0)   & MTDv2               & \textbf{1.216} & \textbf{0.387} & \textbf{0.358} & \textbf{0.132} \\
(d1)   & w/ SPN             & 1.363 & 0.416 & 0.361 & 0.135 \\
(d2)   & w/o Surface modeling residual    & 1.571 & 0.448 & 0.362 & 0.136 \\
(d3)   & w/o Discontinuity density residual       & 1.370 & 0.412 & 0.359 & 0.134 \\
(d4)   & w/o Geometry-guided affinity  & 1.284 & 0.395 & 0.358 & 0.133 \\
(d5)   & w/o Geometry-aware segmentation    & 1.702 & 0.465 & 0.365 & 0.138 \\
\bottomrule
\end{tabular}
\end{table}

First, segment-wise affine transformation is applied only to the qualified local segments. Least-squares fitting (LSF) is used by default to fit the inverse depth data, and several alternative affine transformation parameter estimation strategies are considered for comparison. It can be observed that, when LSF is adopted in both cases, estimating the affine transformation parameters from inverse depth consistently achieves higher accuracy than estimating them from depth. This is primarily because the relative depth estimation network~\cite{depthanythingv2} is originally trained to estimate affine-invariant inverse depth, making its estimates inherently more compatible with affine transformation parameter estimation from inverse depth. LSF is then compared with median alignment based on median statistics, quantile matching that aligns the distributions of estimated and metric depth using selected quantiles, and moment matching that aligns statistical moments such as the mean and variance. The results show that LSF consistently achieves better depth completion performance than other parameter estimation strategies.

Second, the efficacy of sparse graph optimization for estimating the affine transformation parameters of the unqualified local segments is evaluated. A noticeable improvement is observed after incorporating sparse graph optimization based on 2D graph distances. To validate that this improvement arises from the proposed graph-based parameter propagation, an alternative setting is evaluated in which the globally fitted affine transformation parameters are directly applied to the unqualified segments. The results suggest that this alternative leads to a significant performance degradation, further demonstrating the effectiveness of the proposed sparse graph optimization.

Furthermore, introducing the explicit depth refinement process on top of sparse graph optimization yields the default configuration of MTD and results in a substantial improvement. Internal ablations are then conducted on the main components of this depth refinement process. The discontinuity-aware geodesic cost defined in~\eqref{eq:def-dphi} is first replaced with a cost computed directly from the Laplacian of the depth map. This alternative degrades depth completion performance, demonstrating the effectiveness of the proposed discontinuity-aware geodesic cost. The second-order surface modeling for propagated depth estimation defined in~\eqref{eq:local_second_order} is then replaced with B-spline interpolation. This change also leads to a decline in depth completion performance, demonstrating the effectiveness of the proposed surface-modeling-based depth propagation approach. Finally, the maximum number of depth propagation steps, which is set to 3 by default in Sect.~\ref{sect.implementation}, is set to both smaller and larger values. The results show that reducing the maximum number of depth propagation steps to 1 consistently leads to a degradation in depth completion performance, as the limited propagation range prevents sufficient pixels from being influenced by the metric observations. In contrast, increasing the maximum number of steps to 5 further improves depth completion accuracy on the KITTI dataset but leads to performance degradation on the VOID dataset. These results suggest that a larger propagation range has the potential to improve depth refinement performance but may also increase the risk of error accumulation. Furthermore, the optimal propagation range varies with scene structure and the spatial distribution of sparse metric observations across datasets, indicating that manually selected propagation parameters have limited generalizability and further highlighting the importance of the subsequent data-driven depth refinement.

Building upon MTD, incorporating geometric cues into the superpixel segmentation algorithm and further introducing the data-driven depth refinement process yields the default configuration of MTDv2. Compared with MTD, MTDv2 achieves substantial improvements in depth completion accuracy on both the KITTI and VOID datasets, demonstrating the effectiveness of the proposed data-driven depth refinement paradigm. Subsequently, the geometry-guided propagation head within MTDv2 is compared with a conventional spatial propagation network (SPN) adopted in the prior work~\cite{nlspn}, and the results demonstrate the effectiveness of the proposed design. In addition, each internal component of the geometry-guided propagation head is individually removed to validate its effectiveness. The results show that removing the surface modeling residual minimization defined in~\eqref{eq:candidate_residual}, the discontinuity density residual minimization defined in~\eqref{eq:energy_residual}, or the geometric guidance for affinity estimation defined in~\eqref{eq:sda_affinity} consistently degrades depth completion performance compared with the full configuration of MTDv2, demonstrating the contribution of each component. Finally, replacing the geometry-aware superpixel segmentation with the purely texture-based segmentation used in MTD also leads to a notable performance drop compared with MTDv2. This result shows the complementary effects of geometry-aware superpixel segmentation and data-driven depth refinement, where the former reduces the burden on the latter by limiting large cross-segment depth errors, allowing the latter to focus on fine-grained depth refinement, as discussed in Sect.~\ref{sect.segment}.

\begin{figure}[!t]
	\centering
	\includegraphics[width=0.49\textwidth]{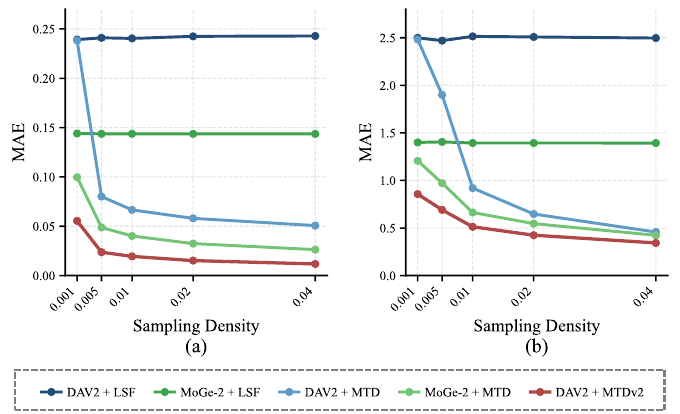}
	\caption{
    Quantitative evaluation of MTDv2's sensitivity to varying sampling densities of metric observations on (a) the indoor subset of the ETH3D dataset and (b) the outdoor subset of the DIODE dataset, compared with MTD and global affine transformation using different foundation models.
    }
\label{fig.sensitivity}
\end{figure}

\begin{figure}[!t]
	\centering
	\includegraphics[width=0.49\textwidth]{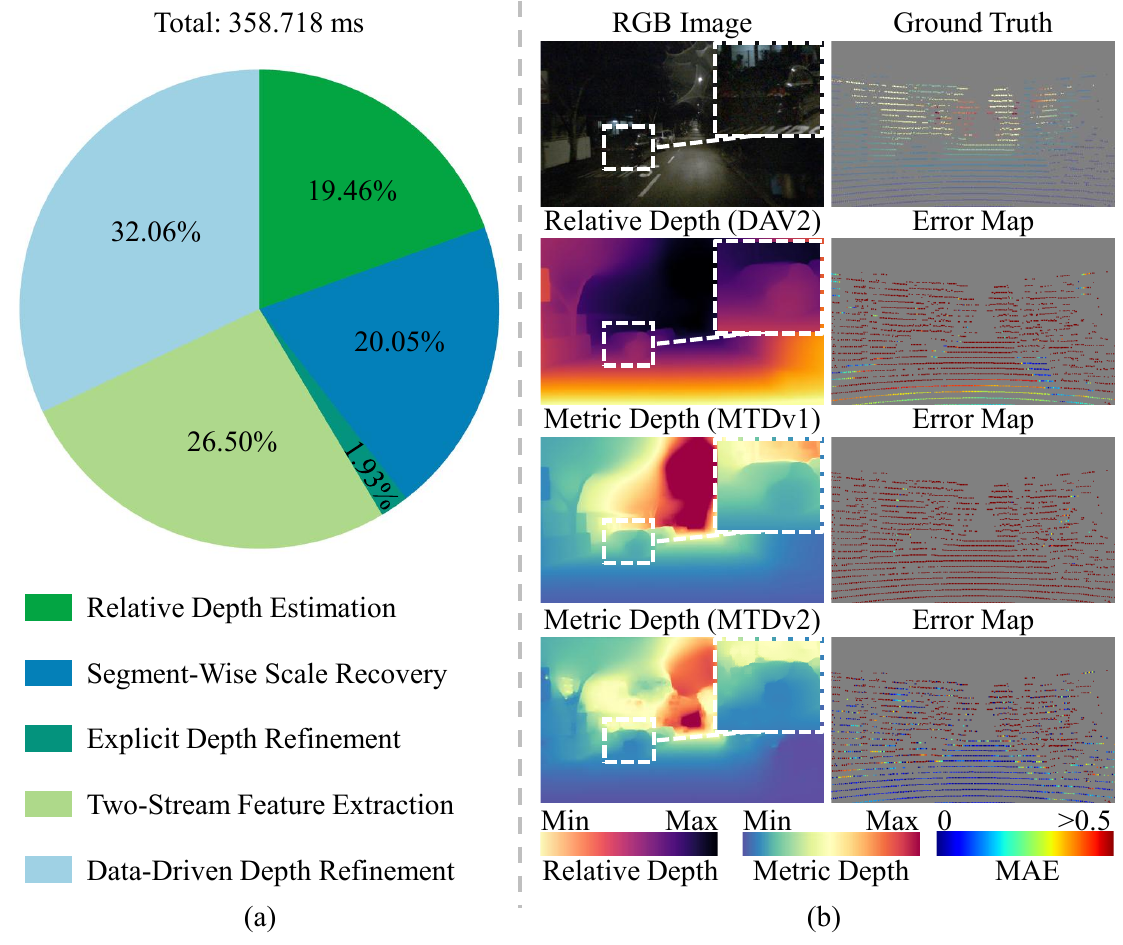}
	\caption{
    Inference time breakdown and a representative failure case of the proposed method under challenging conditions.}
\label{fig.discussion}
\end{figure}

\subsection{Sensitivity Analyses}
\label{sect.efficient}

In practice, only sparse metric depth observations are typically available because LiDAR and radar sensors cannot acquire sufficiently dense measurements within a limited sensing time.
Therefore, this study evaluates MTDv2 under different sampling densities to analyze its sensitivity to observation sparsity.
The experimental results are presented in Fig.~\ref{fig.sensitivity}. First, MTDv2 with Depth~Anything~V2 consistently outperforms MTD with either DAV2 or MoGe-2, even though MoGe-2 itself provides more accurate depth estimates than DAV2.
Moreover, the performance advantage becomes increasingly pronounced as metric observations become sparser. This trend can be attributed to the fact that explicit depth refinement becomes increasingly constrained by the limited metric observations available for depth metricization, making data-driven depth refinement more important under highly sparse observation settings.

\subsection{Discussion}

Despite the compelling results achieved, the proposed framework still has the following three limitations.

First, the computational efficiency of MTDv2 remains limited. A detailed breakdown of the inference time is given in Fig.~\ref{fig.discussion}(a). It can be observed that the relative depth estimation network and the network components used for data-driven depth refinement employ independent backbones for feature extraction, together accounting for approximately $50\%$ of the total inference time. Therefore, a more compact network design with appropriate training strategies could be explored to facilitate effective feature sharing between these components while preserving their respective functions, thereby reducing inference overhead without compromising depth completion performance. In addition, the superpixel segmentation algorithm used for segment-wise scale recovery relies solely on CPU execution, and GPU-accelerated implementations could be explored to further improve computational efficiency. Furthermore, the depth propagation head in the data-driven depth refinement process introduces substantial inference overhead when predicting residuals and propagation logits, motivating the development of a more lightweight architecture.

Second, the proposed method becomes less robust when the relative depth estimates are inaccurate. As shown in Fig.~\ref{fig.discussion}(b), under poor illumination conditions, the estimated relative depth of the car becomes erroneous, and the resulting error is also observed in the outputs of both MTD and MTDv2. This is because the explicit depth refinement process largely preserves the initial geometric structure derived from the relative depth estimation and is therefore incapable of correcting substantial depth errors. Although the data-driven metric depth refinement process can reduce depth residuals, its effectiveness remains limited when similar challenging conditions are insufficiently represented in the annotated training data. Therefore, one possible direction for addressing this issue is to incorporate effective domain adaptation strategies or robust unsupervised learning methods, enabling both the relative depth estimation network and the learnable depth refinement module to better adapt to challenging scenarios by leveraging additional unlabeled data.

Finally, the proposed framework still relies on accurate extrinsic calibration between the camera and range sensor, provided by public datasets. However, in practical use, the actual extrinsic parameters may deviate from their initial values obtained through offline target-based calibration due to sensor mounting variations, mechanical perturbations, or long-term system changes, which can degrade depth metricization accuracy. Integrating the proposed method with high-performance online target-free extrinsic calibration approaches, such as our previous work~\cite{huang2026dst}, could provide a more practical system capable of operating robustly under imperfect calibration conditions.

\section{Conclusion and Future Work}
\label{sect.conclusion}

This article presented MTDv2, a hybrid metric depth recovery framework that combines explicit optimization-based depth metricization with data-driven metric depth refinement. The proposed framework first resolves relative-to-metric ambiguity through segment-wise affine parameter estimation and sparse graph-based parameter propagation, followed by discontinuity-aware dynamic programming to improve local geometric fidelity. Building upon the explicitly metricized depth, the hybrid refinement process further incorporates learnable residual correction and geometry-guided multi-path depth propagation to overcome the limitations of handcrafted surface modeling and propagation strategies.
Extensive experiments on 16 public datasets demonstrate that MTDv2 achieves robust and generalizable metric depth recovery under diverse sparse observation settings, with competitive performance in depth completion and downstream 3D perception tasks.

Future work will focus on improving computational efficiency, enhancing robustness under challenging observation conditions, and developing unified calibration-free depth perception frameworks.

\small
\normalem
\bibliographystyle{IEEEtran}
\bibliography{ref}

\end{document}

%% file: packages.tex
\usepackage[table,xcdraw]{xcolor}
\usepackage{scalerel}
\usepackage{tikz}
\usetikzlibrary{svg.path}

\definecolor{orcidlogocol}{HTML}{A6CE39}
\tikzset{
  orcidlogo/.pic={
    \fill[orcidlogocol] svg{M256,128c0,70.7-57.3,128-128,128C57.3,256,0,198.7,0,128C0,57.3,57.3,0,128,0C198.7,0,256,57.3,256,128z};
    \fill[white] svg{M86.3,186.2H70.9V79.1h15.4v48.4V186.2z}
                 svg{M108.9,79.1h41.6c39.6,0,57,28.3,57,53.6c0,27.5-21.5,53.6-56.8,53.6h-41.8V79.1z M124.3,172.4h24.5c34.9,0,42.9-26.5,42.9-39.7c0-21.5-13.7-39.7-43.7-39.7h-23.7V172.4z}
                 svg{M88.7,56.8c0,5.5-4.5,10.1-10.1,10.1c-5.6,0-10.1-4.6-10.1-10.1c0-5.6,4.5-10.1,10.1-10.1C84.2,46.7,88.7,51.3,88.7,56.8z};
  }
}
\newcommand\orcidicon[1]{\href{https://orcid.org/#1}{\mbox{\scalerel*{
\begin{tikzpicture}[yscale=-1,transform shape]
\pic{orcidlogo};
\end{tikzpicture}
}{|}}}}